\documentclass[fleqn,10pt,twocolumn]{wlscirep}
\usepackage[utf8]{inputenc}
\usepackage[T1]{fontenc}
\usepackage{amsmath}
\usepackage{amssymb}
\usepackage{csquotes}
\usepackage{booktabs}
\usepackage{dblfloatfix}
\usepackage{multirow}
\usepackage{colortbl}
\usepackage{multicol}

\usepackage{subcaption}

\title{A vision-language model and platform for temporally mapping surgery from video}

\author[1,*]{Dani Kiyasseh}
\affil[1]{Halsted AI, Sunnyvale, California, 94086 USA}
\affil[*]{dani@halstedhealth.ai}

\begin{abstract}
Mapping surgery is fundamental to developing operative guidelines and enabling autonomous robotic surgery. Recent advances in artificial intelligence (AI) have shown promise in mapping the behaviour of surgeons from videos, yet current models remain narrow in scope—capturing limited behavioural components within single procedures, and offer limited translational value, as they remain inaccessible to practising surgeons. Here we introduce Halsted, a vision-language model trained on the Halsted Surgical Atlas (HSA)—one of the most comprehensive annotated video libraries grown through an iterative self-labelling framework and encompassing over 650,000 videos across eight surgical specialties. To facilitate benchmarking, we publicly release HSA-27k, a subset of the Halsted Surgical Atlas. Halsted surpasses previous state-of-the-art models in mapping surgical activity while offering greater comprehensiveness and computational efficiency. To bridge the longstanding translational gap of surgical AI, we develop the Halsted web platform (\url{https://halstedhealth.ai/}) to provide surgeons anywhere in the world with the previously-unavailable capability of automatically mapping their own procedures within minutes. By standardizing unstructured surgical video data and making these capabilities directly accessible to surgeons, our work brings surgical AI closer to clinical deployment and helps pave the way toward autonomous robotic surgery.
\end{abstract}

\begin{document}

\flushbottom
\maketitle

\thispagestyle{empty}

Mapping the behaviour of surgeons reveals substantial variability in surgical decision-making \cite{Birkmeyer2013, Stulberg2020}. While some of this variability reflects innocuous stylistic differences, other patterns may conceal problematic behaviours that warrant intervention. Distinguishing among these sources of variability has broad implications for credentialing, granting operating privileges, and teaching autonomous robotic systems how to perform surgery safely. Achieving this requires annotating procedures using taxonomies to categorize surgeon behaviour from operative videos \cite{Volpe2015, Valdis2016, Kiely2015, Sobel2016}. Yet the rapidly increasing volume of surgical video and the inherent complexity of operative procedures \cite{Childers2023} make comprehensive manual annotation infeasible, underscoring the need for scalable, automated approaches to map surgery.

Machine learning offers a promising solution. Early efforts focused on categorizing surgeon behaviour from video and robotic kinematic data \cite{Zia2018, Funke2019}. Building on this foundation, our prior work automated the detection of procedural steps, recognition of surgical behaviours, and assessment of technical proficiency using video alone \cite{Kiyasseh2023, Kiyasseh2023CommsMed, Kiyasseh2023DigMed}. However, existing approaches remain focused almost entirely on algorithmic advancements that target a single task within a single specialty \cite{Yuan2025, Hu2025} and overlook how to translate such advancements into clinical practice. Without the capacity to perform multiple tasks across a wide range of specialties, and an accessible platform through which surgeons can view procedural insights, these methods have limited real-world applicability and fall short of enabling a comprehensive surgical map that spans the millions of procedures performed worldwide each year \cite{Meara2015}.

\begin{figure*}[!h]
    \centering
    \includegraphics[width=1\linewidth]{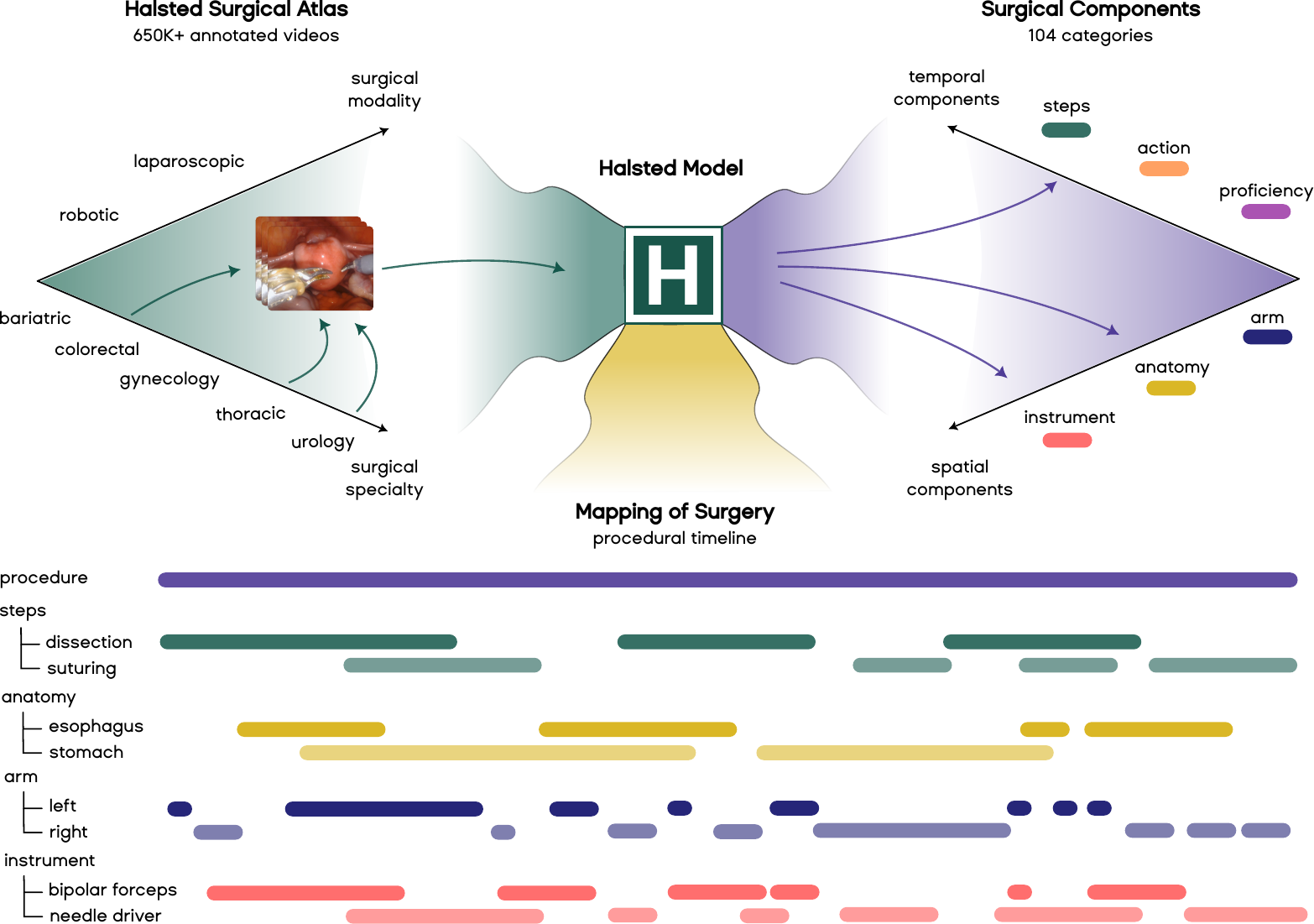}
    \caption{\textbf{Halsted maps surgery from video.} Halsted is trained on the Halsted Surgical Atlas, a library with 650K+ videos, to generate a comprehensive mapping of surgery with 104 categories of surgical components. We present an example of such a mapping for the components of procedure, steps, anatomy, arm, and instrument.}
    \label{fig:main}
\end{figure*}

Here, we introduce Halsted, a vision–language model trained on the Halsted Surgical Atlas, one of the most comprehensive annotated surgical video libraries to date. The atlas comprises over 650K surgical videos spanning 8 specialties and 16 procedures, annotated with 11 surgical components ranging from procedure- and step-level labels to fine-grained actions, anatomy, and technical proficiency, covering 104 distinct categories (Fig.~\ref{fig:main}). To facilitate benchmarking and reproducibility, we open-source HSA-27k, a curated subset of the Halsted Surgical Atlas. As a generative model with outputs controlled by task-specific instructions, Halsted learns to map a wide range of surgical components across temporal scales and outperforms prior state-of-the-art methods on an external benchmark. By jointly training across all tasks and specialties, Halsted implicitly captures relationships between surgical procedures. Finally, to bridge the translational gap, we deploy a web platform powered by Halsted that enables surgeons to automatically map their own procedures within minutes, a previously unavailable capability.

\section*{Results}

\subsection*{Halsted: a vision-language model for temporally mapping surgery}

Mapping surgery exclusively from video presents several unique challenges. First, multi-specialty surgical videos exhibit substantial visual heterogeneity arising from differences in patient anatomy, instrument sets, and surgeon-specific technique. Second, available surgical datasets are often highly imbalanced, with over-representation of common procedures (e.g., robotic prostatectomy) and under-representation of less frequent ones (e.g., robotic cardiac surgery). Third, surgical structure spans a broad range of temporal scales, from low-frequency procedural steps lasting 10–20 minutes to high-frequency atomic actions occurring over seconds. A robust model must therefore effectively handle both spatial and temporal variability while exploiting shared structure across procedures.

Halsted is a vision–language model that performs a temporal mapping of surgery from operative video. It is multimodal, jointly processing visual and textual inputs; multitask, simultaneously predicting multiple surgical components; and generative, producing a structured sequence of tags in an autoregressive manner. The model comprises four core modules: (1) a video encoder that processes short video clips, (2) task embeddings that encode instructional prompts, (3) a tokenizer to encode surgical component tags, and (4) a transformer backbone that autoregressively generates the tag sequence (see Halsted vision language model).

Halsted is trained on the Halsted Surgical Atlas (HSA), the largest video library of its kind, comprising over 650K annotated video clips labeled with surgical component tags. The dataset spans sixteen procedures across eight surgical specialties: bariatric, colorectal, general, gynecology, hepatobiliary, pancreatic, thoracic, and urology. Surgical component tags capture multiple levels of granularity, including procedure (e.g., hysterectomy), steps (e.g., suturing), phases (e.g., needle driving), anatomy (e.g., uterus), instrument (e.g., bipolar forceps), and technical proficiency (see Table~\ref{table:data} for a detailed breakdown). Conditioned on an input video and a task instruction, Halsted generates a task-specific sequence of component tags, enabling a wide range of tasks to be performed within a single unified framework.

\begin{figure*}[!t]
    \centering
    \includegraphics[width=0.95\linewidth]{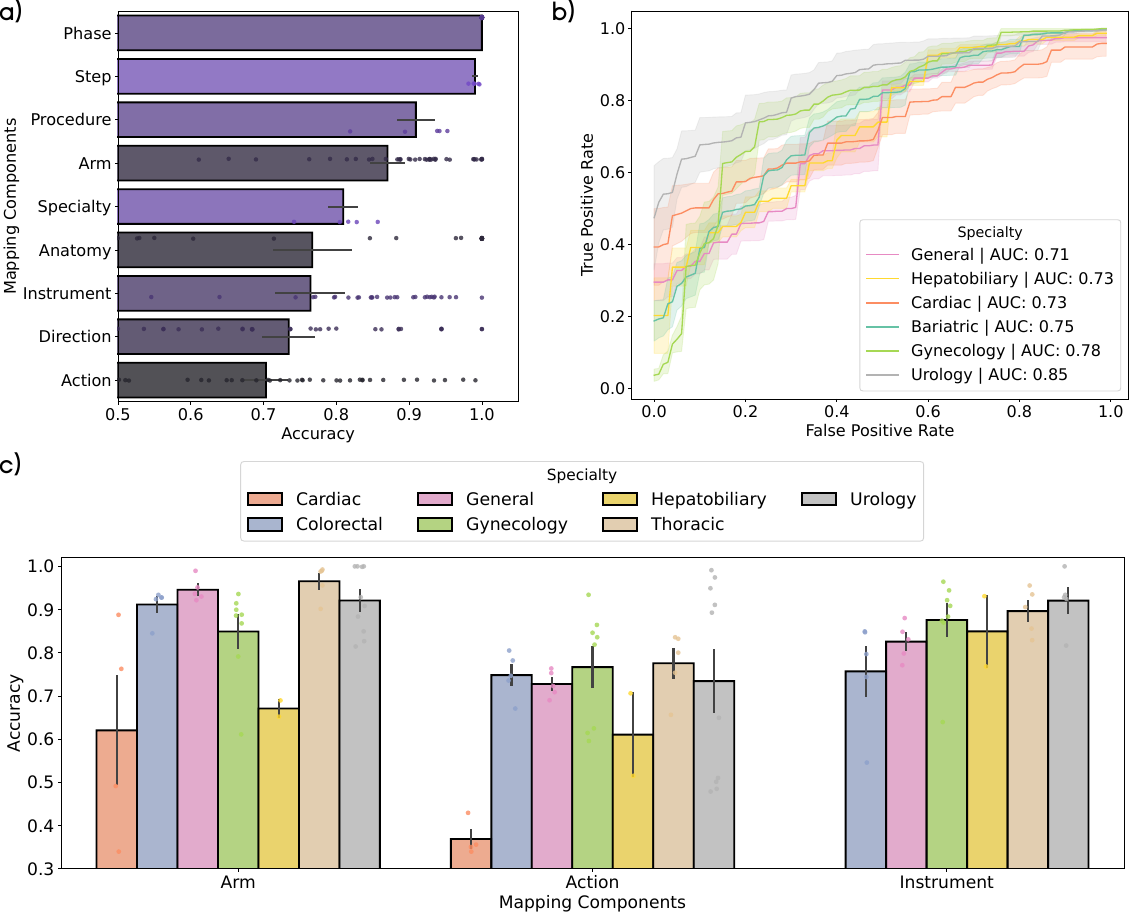}
    \caption{\textbf{Halsted learns to comprehensively map surgery across specialties.} Halsted reliably \textbf{(a)} maps surgical components at various levels of granularity, from anatomy to procedure type, \textbf{(b)} assesses surgical proficiency irrespective of specialty, and \textbf{(c)} recognizes granular surgical components such as arms used, actions performed, and instruments used. The shaded area and error bars reflect one standard error from the mean.}
    \label{fig:halsted_performance}
\end{figure*}

\subsection*{Halsted learns to comprehensively map surgery across temporal scales}

We trained Halsted jointly on all tasks using video clips from all specialties in the Halsted Surgical Atlas (see Training the model). We evaluated Halsted’s ability to map surgical components on held-out sets of videos using 5-fold Monte-Carlo cross-validation, and found that it robustly captures surgical structure across temporal scales (Fig.~\ref{fig:halsted_performance}a). HSA-27k, a subset of the Halsted Surgical Atlas, constitutes the test set for one of these folds (see Evaluating the model and Supplementary Notes 2 and 3).

At the coarsest level, Halsted achieves accuracies of 99\% (95\% CI, 98.2–99.8\%) for identifying surgical steps and 91\% (95\% CI, 84.1–97.8\%) for identifying procedures. At the finest temporal granularity, its accuracy in recognizing stitch direction during suturing and discrete surgical actions is 73.4\% (95\% CI, 66.4–80.5\%) and 70\% (95\% CI, 64.6–70.6\%), respectively. This performance is notable given the large action space, comprising 19 distinct action categories. We further report strong but category—dependent performance in temporally mapping surgical actions using the F1 score \cite{Reinke2025} (see Supplementary Note 3).

Halsted’s comparatively lower performance on the cardiac specialty is expected, as the model was not exposed to cardiac videos during training. Only two IMA harvest videos were available and were reserved exclusively for validation and testing (see Table~\ref{table:data} and Evaluating the model in Methods).

\subsection*{Halsted assesses suturing technical proficiency across surgical specialties}

We conditioned the same Halsted model to assess binary technical proficiency for two suturing activities—needle handling and needle driving—based on established assessment criteria \cite{Haque2022} (Fig.~\ref{fig:halsted_performance}b). Halsted is able to assess technical proficiency, though performance varies across specialties, achieving an AUROC of 0.71 (95\% CI, 0.68–0.74) in general surgery and an AUROC of 0.85 (95\% CI, 0.78–0.92) in urology. These results indicate that, despite the long-standing assumption that technical proficiency requires specialty-specific assessment rubrics, suturing exhibits sufficient cross-specialty common structure for Halsted to learn and generalize from during training.

\begin{figure*}[!h]
    \centering
    \includegraphics[width=1\linewidth]{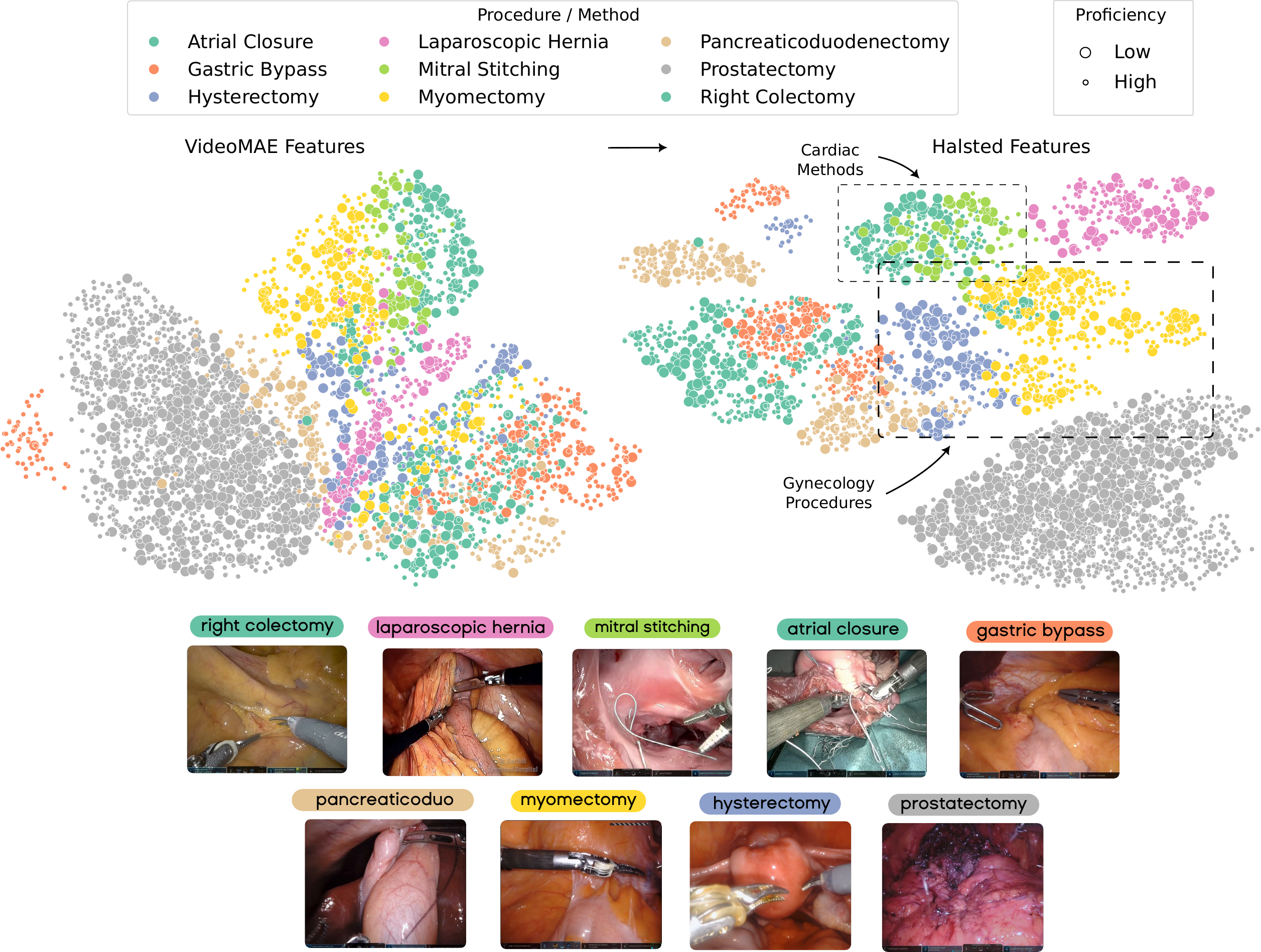}
    \caption{\textbf{Halsted learns a nuanced relationship between surgical videos.} We present the two-dimensional UMAP embeddings of representations of video clips extracted by VideoMAE (left) and of the features generated by Halsted (right) in the final layer of the transformer decoder when tasked with assessing suturing proficiency. Each colour reflects a distinct procedure and marker size indicates proficiency (large markers indicate low proficiency). Although VideoMAE can distinguish between procedures, Halsted takes this one step further and clusters procedures of the same specialty, as observed with cardiac and gynecology procedures.}
    \label{fig:halsted_embeddings}
\end{figure*}

\subsection*{Halsted learns the similarity of surgical procedures without explicit programming}

In mapping surgery, Halsted leverages a powerful pre-trained vision encoder (VideoMAE \cite{Tong2022}). To illustrate this, we compute a two-dimensional UMAP \cite{Mcinnes2018} embedding of representations extracted by VideoMAE from video clips in the held-out test set of the proficiency task (Fig.~\ref{fig:halsted_platform}, left). These representations reasonably separate surgical procedures, albeit capturing only one facet of surgical structure, but already provide Halsted with a strong initialization for downstream surgical mapping.

We next embed representations from the final decoder layer of Halsted after joint training on video clips from all procedures (Fig.~\ref{fig:halsted_platform}, right). Compared with VideoMAE alone, Halsted’s learned representations exhibit improved separability across procedures and, importantly, reveal implicit relationships between procedures, despite the model never being explicitly programmed with procedural hierarchies or similarities. For example, Halsted clusters the cardiac methods of atrial closure and mitral stitching, as well as the gynecologic procedures hysterectomy and myomectomy, reflecting their intra-specialty similarity. Although not a groundbreaking finding, this emergent organization indicates that Halsted learns meaningful surgical structure and context, an essential property for a model designed to map surgery holistically.

\subsection*{Halsted outperforms state-of-the-art surgical computer vision model} 

To address new tasks, prior state-of-the-art models in surgical video analysis typically require either retraining all model parameters or introducing additional task-specific parameters, approaches that are impractical at scale (see Halsted is designed for computational efficiency and maintainability). Halsted represents a departure from this paradigm by mapping a wide range of surgical components using a single shared set of parameters, with task behaviour specified through instructional conditioning.

In addition to these architectural advantages, we benchmark Halsted against a state-of-the-art model, SAIS \cite{Kiyasseh2023}, which has previously been shown to map multiple surgical components but requires separate parameterizations per task (see Benchmarking against state-of-the-art model). We find that SAIS struggles to scale to tasks with large category counts and to datasets spanning multiple surgical specialties. To enable a fair comparison, we therefore train SAIS on a simplified setting limited to suturing activity recognition (3 categories) and technical proficiency assessment (2 categories).

Under this constrained setup, Halsted substantially outperforms SAIS, achieving an AUROC of 1.00 (95\% CI, 1.00–1.00) versus 0.68 (95\% CI, 0.65–0.72) for suturing activity recognition, and an AUROC of 0.78 (95\% CI, 0.73–0.77) versus 0.73 (95\% CI, 0.70–0.74) for technical proficiency assessment. Beyond raw performance, a key limitation of SAIS is that separate models are required for each task, with costs in training, maintenance, and deployment—implications we discuss in a subsequent section.

\subsection*{Halsted's lightweight decoder suffices for mapping surgery}

Motivated by empirical evidence that larger models with greater parameter counts often outperform smaller architectures, we investigated whether increasing decoder capacity improves Halsted’s performance. Specifically, we performed a drop-in replacement of the original 2-layer transformer decoder (44M parameters) with a pre-trained Llama-3.2 decoder (1B parameters), and fine-tuned the full model on the micro-activity recognition task in the Halsted Surgical Atlas (see Investigating the effect of decoder size).

\begin{figure}[!t]
    \centering
    \includegraphics[width=0.9\linewidth]{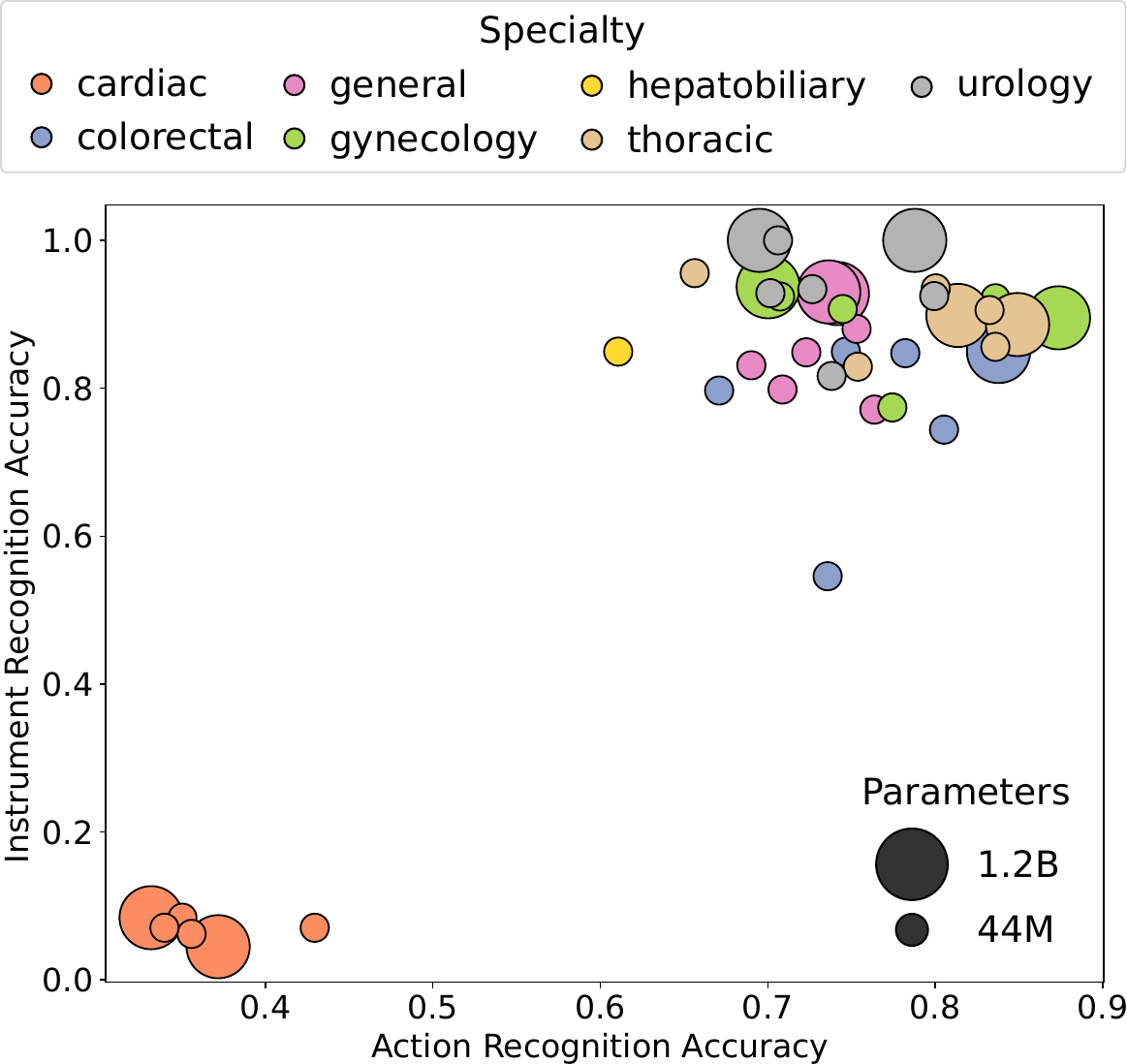}
    \caption{\textbf{Halsted's performance in mapping micro-activity as a function of decoder size.} We train Halsted with a 2-layer decoder or a Llama-3.2 (1B) decoder to perform the micro-activity task, jointly mapping surgical actions and instruments. We report performance using 5-fold Monte-Carlo cross-validation and show that these models perform on par with one another, irrespective of the size of the decoder.}
    \label{fig:size_effect}
\end{figure}

We find that the lightweight Halsted model performed comparably to its Llama-based counterpart despite a ~30× reduction in decoder parameters (see Fig.~\ref{fig:size_effect}). For surgical action recognition, the lightweight and Llama-based models achieved accuracies of 70.3\% (95\% CI, 64.6–76.1\%) and 70.9\% (95\% CI, 68.0–97.0\%), respectively (p > 0.15). Similarly, for instrument-use recognition, accuracies were 76.4\% (95\% CI, 67.0–86.0\%) and 81.4\% (95\% CI, 62.8–99.9\%), respectively (p>0.15). These findings challenge the commonly held assumption that larger models necessarily yield superior performance.

Even if marginal gains were observed with a substantially larger decoder, such improvements would need to be weighed against the significantly increased memory footprint and computational cost (see Halsted is designed for computational efficiency). We attribute Halsted’s strong performance to the high-quality initialization provided by the vision encoder (Fig.~\ref{fig:halsted_embeddings}), which enables a lightweight decoder to be sufficient for the targeted surgical understanding tasks.

\subsection*{Validating Halsted on external dataset}

To assess whether Halsted’s robust performance generalizes beyond the Halsted Surgical Atlas, we evaluated its ability to map a single surgical component—actions—on videos from the publicly available RARP-50 benchmark dataset. Halsted was trained on the provided training split and evaluated on the held-out test set. Owing to differences in evaluation protocols across competing methods (see External validation of Halsted), we report performance relative to a random-chance classifier (Table~\ref{table:external_validation}), defined as predicting an action uniformly at random. Under this evaluation, Halsted achieves the largest relative improvement in accuracy—a $5.5\times$ gain over chance—when tasked with recognizing all eight discrete action categories in the test set.

\begin{table}[!b]
    \centering
    \begin{tabular}{c|c c c}
        \toprule
        & \multicolumn{2}{c}{Accuracy (\%)} & Relative \\
         Method & Random & Actual & Improvement \\
         \midrule
         MA-TCN \cite{Van2022} & 25.9 &  80.9 & 3.1$\times$\\
         SAIS \cite{Kiyasseh2023} & 14.3 & 59.8 & 4.2$\times$ \\
         Halsted & 12.5 & 68.6 & \textbf{5.5$\times$} \\
         \bottomrule
    \end{tabular}
    \caption{\textbf{Halsted performance on RARP-50 benchmark.} We compare the actual accuracy of the model to that of random chance (i.e., randomly guessing an action category) and report the relative improvement. Previous results are borrowed from previous work \cite{Kiyasseh2023}. We show that Halsted achieves the greatest improvement when identifying all eight action categories on the RARP-50 test set.}
    \label{table:external_validation}
\end{table}

\begin{figure}[!b]
    \centering
    \includegraphics[width=0.9\linewidth]{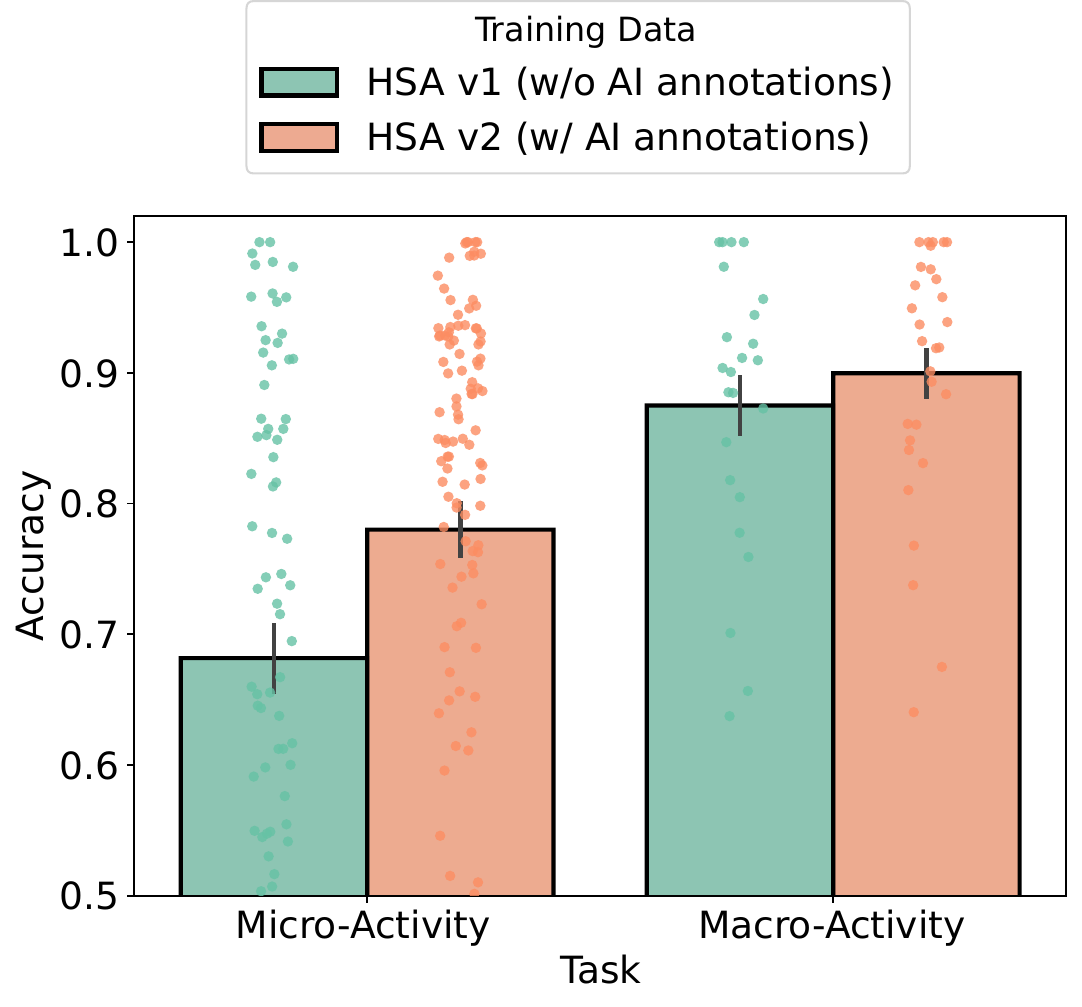}
    \caption{\textbf{Halsted's performance improves with a self-learning strategy.} We train Halsted on HSA v1, the first version of the Halsted Surgical Atlas without any AI-generated annotations, and HSA v2, the final version of the Halsted Surgical Atlas after incorporating AI-generated annotations. For details on the iterative labelling process, see Methods. Both models are evaluated on the same held-out test sets using 5-fold Monte-Carlo cross-validation.}
    \label{fig:hsa_effect}
\end{figure}

\subsection*{Investigating the effect of AI-generated surgical component tags}

The development of Halsted followed an iterative training and data expansion lifecycle. We first trained an initial version of the model on manually-annotated video clips from HSA v1 (see Table~\ref{table:data} for the distribution of manual annotations). This preliminary model was then deployed on full-length, unlabelled surgical videos to generate AI-derived surgical component tags at the clip level (see Iterative development of the Halsted Surgical Atlas). This self-labelling strategy expanded the Halsted Surgical Atlas by approximately four-fold, yielding HSA v2, and provided substantially broader data coverage for training the second version of Halsted (a single round of self-labelling was performed).

Despite incorporating multiple quality assurance steps, we explicitly evaluated the fidelity of AI-generated annotations using the following logic. If Halsted were to achieve comparable performance regardless of annotation source, this would suggest that AI-generated labels are similar in quality and reliability to manually curated annotations. Focusing on the macro-activity task, for which a subset of annotations was AI-generated, we observe no statistically significant difference in performance on video clips with different annotation sources (p = 0.25, Wilcoxon signed-rank test). Specifically, Halsted achieves an accuracy of 89\% (95\% CI, 81–93\%) on manually annotated clips and 96\% (95\% CI, 88–97\%) on clips with AI-generated annotations.

\begin{figure*}[!t]
    \centering
    \includegraphics[width=1\linewidth]{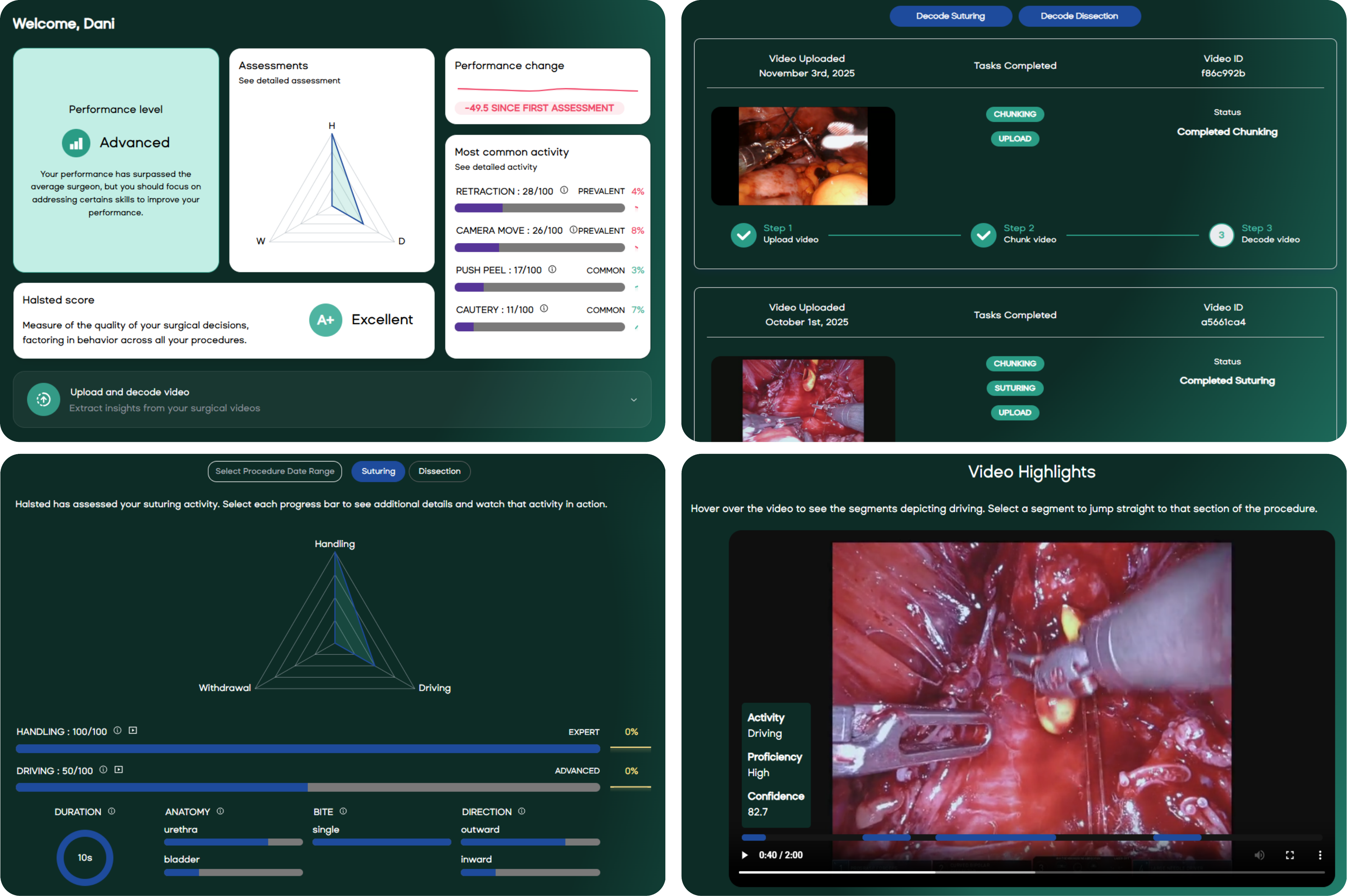}
    \caption{\textbf{Overview of the Halsted Platform.} The Halsted Platform is a secure web-based platform that enables surgeons to upload a surgical video and receive an automated mapping of their surgical activity within minutes. Upon logging in, surgeons are presented with a high-level summary of their surgical activity. They have the flexibility to select the mapping task (e.g., micro-activity or proficiency) before viewing a detailed breakdown of their activity. Surgeons can also better contextualize this activity through visual overlays pinpointing exactly when the activity occurred during the procedure.}
    \label{fig:halsted_platform}
\end{figure*}

\subsection*{Halsted benefits from a self-learning strategy}

Through a self-labelling strategy, we expanded the Halsted Surgical Atlas by approximately four-fold prior to training the final Halsted model (see Iterative development of the Halsted Surgical Atlas in Methods). Although increasing dataset size often improves model performance, such gains are not guaranteed—particularly when newly added data may contain noisier annotations. We therefore quantified the impact of this expanded video library on Halsted’s performance.

Specifically, we compared Halsted’s performance on held-out test sets when trained exclusively on manually curated annotations versus when trained on the full dataset, including video clips with AI-generated annotations (Fig.~\ref{fig:hsa_effect}). Incorporating AI-labelled clips into training yields a statistically significant improvement in performance on the micro-activity task, increasing accuracy from 68\% (95\% CI, 63.1–73.2\%) to 78\% (95\% CI, 74.0–82.0\%; p < 0.05). Performance on the macro-activity task also improves, from 88\% (95\% CI, 83.3–92.0\%) to 90\% (95\% CI, 86.3–93.6\%).

\subsection*{Halsted web platform enables surgeons anywhere to map their own procedures}

Until now, surgeons have either lacked access to any tool capable of comprehensively mapping surgical procedures or have been forced to manually annotate their own videos, an unrealistic expectation given their clinical workload. Prior academic work has largely centred on algorithmic innovation without translating these methods into tools that practising surgeons can use, resulting in limited real-world impact. We begin to close this translational gap by going beyond algorithm development and delivering a secure and scalable web platform that enables surgeons—regardless of geographical location—to analyse and map their own procedures (Fig.~\ref{fig:halsted_platform}).

Upon logging into the platform, surgeons are immediately shown a personalized summary of their past performance (Fig.~\ref{fig:halsted_platform} upper left), automatically generated from their uploaded operative videos. The platform includes an organized library of all video uploads (Fig.~\ref{fig:halsted_platform} upper right), and allows surgeons, with a single click, to comprehensive map specific surgical activity such as suturing. Conceptually, this is equivalent to supplying the model with the instructions defined in Task 3 (see Fig.~\ref{fig:halsted}). For a detailed description of the underlying video-processing pipeline, see the Halsted platform section.

Surgeons can now review their suturing activity in depth, including the proportion of stitches executed with high proficiency, the duration of each phase of the stitch, the anatomical targets being sutured, and more (Fig.~\ref{fig:halsted_platform} lower left). Automatically extracting such rich and structured information from video would not have been feasible without the Halsted model. It has previously been demonstrated that such metrics are predictive of postoperative patient outcomes \cite{Kiyasseh2023, Heard2025}. Further, we enable surgeons to better understand the temporal context in which their activity was performed by overlaying our AI-based temporal annotations on the corresponding surgical video, pinpointing exactly when a particular event has occurred during the procedure (Fig.~\ref{fig:halsted_platform} lower right).

Together, these capabilities, none of which have previously existed for surgeons, can bring surgical AI closer to clinical deployment. To maximize accessibility and impact, we offer a free version of the Halsted platform to all surgeons, giving them unprecedented access to objective surgical analytics.

\subsection*{Halsted is designed for computational efficiency and maintainability}

The majority of computer vision models for surgical video analysis focus on a single task (e.g., assessing technical proficiency). Adapting such models to additional tasks typically requires either training a separate model with a new set of parameters or adding task-specific parameters to the existing model. Both approaches are impractical for several reasons. First, training models from scratch is computationally expensive and time-consuming. Second, maintaining a suite of task- and specialty-specific models imposes a significant development and deployment burden, increasing the risk of errors. Third, the memory, compute, and energy constraints of edge devices (e.g., GPUs embedded in surgical robots) limit the number of models that can be deployed. For example, deploying separate models for four tasks across eight specialties (4 × 8 = 32 models) would require roughly 38 GB of memory, occupying more than half of the total built-in storage of today’s high-end edge devices. Such approaches slow the deployment of new models, ultimately impacting the surgeon experience.

Halsted addresses these challenges by being a conditional generative model jointly trained across all tasks and specialties. It requires only 1.2 GB of memory, representing a ~30× improvement in memory efficiency over the conventional approach. On an A4000 GPU with mixed-precision inference, Halsted processes a 10-second video clip (16 frames) in 0.31 s, corresponding to roughly 50 frames per second. On the web platform, where videos are processed offline after a procedure, a multi-stage workflow generates a comprehensive surgical map (see Methods), incorporating sequential forward passes and filtering steps to ensure high-quality outputs; this workflow averages 15 minutes to map a 1-hour video. Halsted’s efficient architecture and low latency also reduce cloud-deployment costs, where pricing scales with both memory and compute usage.

\section*{Discussion}

The main contributions of our work are threefold, spanning data, modelling, and infrastructure. We curated the Halsted Surgical Atlas (HSA), a large-scale annotated video library covering eight surgical specialties and 16 procedures, with annotations spanning four tasks, 11 surgical component tags, and 104 distinct categories. To expand the dataset efficiently, we introduced a self-labeling strategy, enabling HSA to grow naturally as new surgical videos become available. For benchmarking and reproducibility, we publicly release HSA-27k, a subset of the atlas. Leveraging this dataset, we developed Halsted, a multimodal, multitask, generative vision-language model that maps surgery exclusively from video. By conditioning Halsted on task instructions—without any modification of its parameters—it can comprehensively map surgical workflows across temporal scales, from identifying steps and assessing technical proficiency to recognizing fine-grained actions. Halsted also implicitly learns relationships between procedures, a critical feature for holistic surgical understanding. Compared with SAIS, a previous state-of-the-art model, Halsted demonstrates superior performance on both HSA and an external benchmark, while offering practical advantages: it is lightweight, controllable, and capable of handling multiple tasks without re-training, making it suitable for low-latency edge and cloud deployment. To facilitate translation into practice, we developed the Halsted web platform, designed to deliver automated surgical insights directly to surgeons. Accessibility remains a limiting factor for the utility of state-of-the-art models; by providing free access, we bridge the gap between advanced AI models and clinical end-users.

The lack of large, annotated, multi-specialty surgical video datasets has historically hindered the development of surgical vision-language models \cite{Ye2025, Derathe2025, Carstens2023, Nwoye2022activity, Goodman2024, Ghamsarian2024}. HSA addresses this gap, enabling a single model trained jointly across all tasks and specialties to learn from a broader, more diverse distribution while exploiting shared structure and semantics across surgical domains. In developing Halsted, we draw on concepts from multimodal language modeling \cite{Wang2023, Song2024}, including autoregressive objectives for vision tasks \cite{Yang2023}, instruction-tuning \cite{Liu2024}, and unified models capable of solving multiple tasks simultaneously \cite{Lu2024}. Our approach is inspired by Pix2Seq \cite{Chen2022}, where multiple vision tasks are addressed by generating a sequence of structured outputs; in our case, these outputs correspond to pre-defined surgical component tags. Importantly, Halsted is orthogonal to recent efforts in self-supervised pretraining of surgical vision encoders \cite{Wei2025, Schmidgall2024, Yang2025, Jaspers2025, Che2025}, which can serve as modular components within our architecture (Fig.~\ref{fig:halsted}). Halsted is trained end-to-end on annotated data from the start, enabling it to map surgery immediately without additional fine-tuning.

Prior work in surgical mapping has largely focused on algorithmic innovations, with limited attention to bridging the translational gap. Many models have been developed to identify isolated surgical components—such as technical proficiency—across various data modalities, including surgical video. However, without an accessible platform, these insights remain effectively inaccessible to the end users: surgeons themselves. We address this translational gap by developing a secure web platform that enables surgeons to automatically map their own procedures within minutes. Surgeons now have direct access to the Halsted model, and as the model evolves, they continue to benefit from its improved capabilities. While the Halsted platform is a commercial product, we provide a free version to support broad adoption. Previous attempts to evaluate surgical AI in applied settings—for instance, in performance feedback \cite{Fazlollahi2022, Giglio2025}—have been limited to one-time trials, confined to laboratory or virtual reality environments, or restricted to a single specialty or task, such as proficiency assessment. In contrast, the Halsted platform represents a step change in how surgeons interact with their live surgical video data, offering unprecedented accessibility.

In mapping surgery, Halsted unlocks the latent value of surgical videos, most of which are either discarded shortly after procedures or archived without review. For surgeons who already use video for performance assessment or education \cite{Schlick2020, Yanik2024, Makary2013, Boyle2025}, Halsted provides a standardized representation of surgical content, enabling systematic comparisons across procedures. More broadly, comprehensive surgical mapping facilitates quantitative analyses of procedural variability and its relationship to patient outcomes \cite{Heard2025}, informing the development of future operative guidelines.

We acknowledge several limitations of our work. First, despite being the most comprehensive dataset of its kind, the Halsted Surgical Atlas does not yet include videos from all surgical specialties and procedures. In particular, by not training on annotated videos from neurosurgery, orthopaedics, and plastic and reconstructive surgery, Halsted’s ability to map procedures in these domains is limited, reducing its utility for surgeons in these specialties. We aim to incorporate such videos in future versions of Halsted as they become available. Second, our annotations focus on a subset of visual taxonomies, a choice guided by prior work \cite{Kiyasseh2023} and the relevance of these taxonomies for predicting patient outcomes \cite{Heard2025}. Nonetheless, Halsted is designed to be flexible and can accommodate alternative or specialty-specific taxonomies as they are developed. Third, the quality of surgical component annotations remains a potential limitation. While annotations are not always perfect, they can still be sufficiently informative for model training. We employed a self-labeling strategy, training Halsted on video clips annotated both manually and via AI-generated labels. Despite a series of quality assurance steps applied before integrating AI-annotated clips into the atlas, a subset of the training data may be incorrect. Consequently, Halsted may occasionally produce erroneous surgical mappings, a limitation that surgeons must be fully aware of when interpreting outputs.

Overall, our findings provide momentum for further exploration of surgical vision-language models and their translational impact on surgical practice. Future work will focus on quantifying how mapped surgical components, and their integration within the Halsted platform, affect surgeon behaviour, patient outcomes, and the education of the next generation of surgeons through prospective clinical trials.

\section*{Methods}

\subsection*{Mapping surgery is the overarching goal}
\begin{figure*}[!h]
    \centering
    \includegraphics[width=1\linewidth]{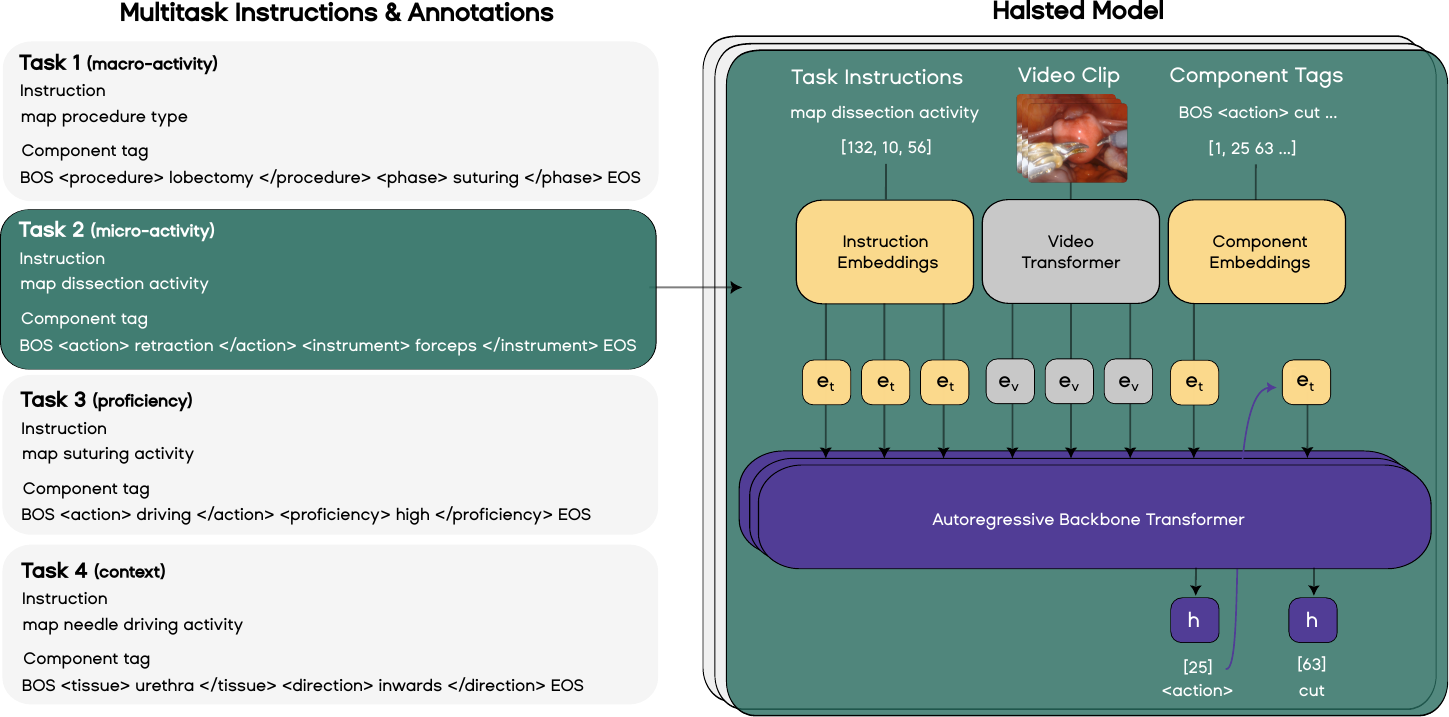}
    \caption{\textbf{Overview of the Halsted model.} Halsted is conditioned on a video clip and a set of task instructions to generate a sequence of surgical component tags. It is a multimodal, multitask, and generative model whose outputs can be controlled via different task instructions. A transformer decoder is presented with a sequence of task instruction embeddings and a video clip embedding to autoregressively generate a sequence of surgical component tags. During inference, we perform greedy decoding at each time-step to select the most likely token (e.g., [25] which corresponds to the word <action>) and retrieve its embedding before feeding it into the model at the next time-step.}
    \label{fig:halsted}
\end{figure*}

Our goal is to comprehensively map a surgery exclusively from video. To achieve this, we develop the Halsted model (Fig.~\ref{fig:halsted}) to generate a sequence of surgical component tags when conditioned on a video clip and a set of instructions. 

\subsection*{Halsted vision-language model}

The Halsted vision-language model is composed of four main modules: 1) a video encoder to process a video clip, 2) task embeddings to process a set of instructions, 3) a tokenizer to process surgical component tags, and 4) a transformer backbone to autoregressively generate the sequence of component tags. We outline each of these modules in depth.

\textbf{Video encoder to process video clips.} Given a video $v \in R^{T \times C \times H \times W}$ with $T$ frames, $C$ channels, height $H$, and width $W$, we extract features by using an encoder pre-trained on publicly-available videos \cite{Tong2022}. The encoder chunks a video into $M$ cubes $\{v_{i} \in R^{\frac{T}{M_{T}} \times C \times \frac{H}{M_{H}} \times \frac{W}{M_{W}}} \}_{i=1}^{M}$ where $M = M_{T} \times M_{H} \times M_{W}$ is the total number of cubes along the temporal and spatial dimensions. Each cube $v_{i}$ is mapped to an embedding, $e_{i} \in R^{D}$. We average these embeddings to obtain a single video embedding $e_{v} \in R^{D}$. While training the model, we choose not to update the parameters of the video encoder to avoid interfering with its strong inductive bias. 

\textbf{Embeddings to process task instructions.} We use a word-level tokenizer \cite{Kudo2018} to split an instruction $I=$ "map dissection activity" into the words $t_{1} = \mathrm{map}$, $t_{2} = \mathrm{dissection}$, and $t_{3} = \mathrm{activity}$, each of which is a token, $t$. We tokenize all instructions $\{ I_{i} \}_{i=1}^{N}$ in a training dataset of $N$ instructions to form a vocabulary $\mathrm{V}_{I} \in \{ t_{i} \}_{i=1}^{N_{I}}$ with $N_{I}$ unique instruction tokens. Each token is associated with an embedding, $e_{t} \in R^{D}$, that is randomly-initialized and optimized alongside the remaining model parameters. Given a sample instruction with $S$ tokens $\{ t_{i} \}_{i=1}^{S}$ tokens, we generate a sequence of instruction embeddings $\{ e_{t_{i}} \}_{i=1}^{S} \in R^{S \times D}$.

\textbf{Tokenizer to process surgical component tags.} We also use a word-level tokenizer to tokenize surgical component tags $\{ A_{i} \}_{i=1}^{N}$ in our \textit{training} dataset of $N$ samples to form a vocabulary $\mathrm{V}_{A} \in \{ t_{i} \}_{i=1}^{N_{A}}$ with $N_{A}$ unique annotation tokens. Each token is associated with an embedding, $e_{t} \in R^{D}$, that is randomly-initialized and optimized alongside the remaining parameters of the model. Notably, we combine all tokens across the vocabularies to form a single vocabulary $\mathrm{V} = \mathrm{V}_{I} \cup \mathrm{V}_{A}$. Given sample surgical component tags with $S$ tokens, $\{ t_{i} \}_{t=1}^{S}$, we generate a sequence of component embeddings, $ \{ e_{t_{i}} \}_{i=1}^{S} \in R^{S \times D}$.

\textbf{Transformer backbone to generate surgical component tags.} We use a transformer composed of $L$ layers to autoregressively generate a sequence of surgical component tags. We concatenate a video embedding $e_{v}$ with a sequence of task instruction embeddings $\{ e_{s} \}_{s=1}^{M}$ along the sequence dimension to create the input prefix to the model $[ e_{v} |e_{s} ] \in R^{(M+1) \times D}$. The sequence of surgical component tag embeddings are subsequently presented to the backbone. We always attend to all tokens in the prefix and adopt causal self-attention over the surgical component embeddings, akin to the approach adopted in previous work \cite{Chen2022}.

\subsection*{Training the model}

We form a dataset of triplets $\mathcal{D} = \{ v, I_{k}, A \}_{i=1}^{N}$ where each sample consists of a video clip, $v$, instructions, $I_{k}$, and surgical component tags, $A$. To enable the model to achieve any of the outlined $K$ tasks, we train it on all tasks simultaneously.

\textbf{Optimizing a mathematical objective function} We teach the model to ingest a video $v$ and an instruction $I$ to generate a sequence of surgical component tokens $\{ t_{i} \}_{i=1}^{S}$. To achieve this goal, we use stochastic gradient descent with a mini-batch of $B$ samples and minimize the next-token-prediction objection function (equation~\ref{eq:ntp_loss}), where the model autoregessively predicts the next surgical component token given all previous tokens, the video, and the instructions. The main appeal of such a generative formulation is that if we wanted to extend the sequence of surgical component tags in the future (e.g., to reflect additional information), we can trivially do so without modifying the model architecture or the training paradigm.

\textbf{Training on multiple tasks simultaneously.}
We jointly optimize all model parameters (task instruction embeddings, component tag embeddings, and backbone parameters) while training on all $K$ tasks simultaneously. This ensures the model is capable of achieving any of the outlined tasks during inference. To focus the model's attention on surgical component generation, we only calculate the loss on the generated component tags.
\begin{equation}
    \mathcal{L} =  - \sum_{b=1}^{B} \sum_{i=1}^{S} \log  p \left( t_{i} | t_{<i}, v, I \right) 
    \label{eq:ntp_loss}
\end{equation}

\textbf{Implementation details.} We train Halsted for 40 epochs on an A4000 GPU with a batch size $B=128$ and perform a global optimization over all tasks, only saving model parameters when the worst-performing task outperforms the previously-saved metric (e.g., AUROC). This ensures we optimize for all tasks simultaneously and prevents us from saving a model that performs exceedingly well on one task at the expense of another. We use a cosine decay with a linear warmup for 2 epochs starting at $0.1 \times lr_{base}$ where the $lr_{base} = 1e^{-4}$. The optimizer we use is AdamW with a weight decay of $1e^{-2}$. The model is trained in FP32 full precision. Each video was split into cubes of $2 \times 14 \times 14$ with a Vision Transformer with $40$ layers. We use $M=8$ embeddings to represent each task instruction and $D=1408$ as the dimension of all embeddings. The autoregressive backbone has $L=2$ layers where each self-attention layer has $8$ heads. We found that scaling the number of task instruction embeddings and the number of self-attention heads with the size of the dataset was critical to the training process. 

\begin{table*}[!t]
    \centering
    \begin{tabular}{l | c | c c c c | c | c c}
        \toprule
        & & \multicolumn{4}{c}{\textbf{Samples}} & & \textbf{Annotations} \\
        \textbf{Surgical Procedure / Method} & \textbf{Videos} & \textbf{Task 1} & \textbf{Task 2} & \textbf{Task 3} & \textbf{Task 4} & \textbf{Total} & \textbf{Manual} \\
        \midrule

        \multicolumn{8}{>{\columncolor{gray!15}}l}{\textbf{Bariatric}} \\
        \midrule
        Gastric Bypass & 5 & 303 & - & 303 & 152 & 758 & 100\% \\
        \midrule

        \multicolumn{8}{>{\columncolor{gray!15}}l}{\textbf{Cardiac}} \\
        Atrial Closure & 6 & 219 & - & 219 & 73 & 511 & 100\% \\
        IMA Harvest & 2 & 370 & 370 & - & - & 740 & 100\% \\
        Mitral Stitching & 6 & 89 & - & 89 & 30 & 208 & 100\% \\
        \midrule

        \multicolumn{8}{>{\columncolor{gray!15}}l}{\textbf{Colorectal}} \\
        Right Colectomy & 7 & 386 & - & 386 & 181 & 953 & 100\% \\
        Total Mesorectal Excision & 13 & 18153 & 18153 & - & - & 36306 & 2.3\% \\
        \midrule

        \multicolumn{8}{>{\columncolor{gray!15}}l}{\textbf{General}} \\
        Laparoscopic Cholecystectomy & 20 & 12950 & 12950 & - & - & 25900 & 2.5\%\\
        Laparoscopic Hernia & 23 & 23238 & 23035 & 203 & 68 & 46544 & 3.8\% \\
        \midrule

        \multicolumn{8}{>{\columncolor{gray!15}}l}{\textbf{Gynecology}} \\
        Endometriosis & 15 & 35428 & 35428 & - & - & 70856 & 1.4\% \\
        Hysterectomy & 8 & 2404 & 2171 & 233 & 115 & 4923 & 21.3\%\\
        Myomectomy & 7 & 388 & - & 388 & 195 & 971 & 100\% \\
        \midrule

        \multicolumn{8}{>{\columncolor{gray!15}}l}{\textbf{Hepatobiliary}} \\
        Pancreaticoduodenectomy & 5 & 7824 & 7507 & 317 & 163 & 15811 & 18.8\%\\
        \midrule

        \multicolumn{8}{>{\columncolor{gray!15}}l}{\textbf{Thoracic}} \\
        Right Middle Lobectomy & 4 & 2151 & 2151 & - & - & 4302 & 58.5\%\\
        Right Upper Lobectomy & 9 & 4490 & 4490 & - & - & 8980 & 5.1\% \\
        Segmentectomy & 17 & 12371 & 12371 & - & - & 24742 & 6.1\% \\
        \midrule

        \multicolumn{8}{>{\columncolor{gray!15}}l}{\textbf{Urology}} \\
        Prostatectomy & 469 & 208525 & 205046 & 1497 & 695 & 415763 & 36.6\% \\
        \midrule
        
        \textbf{Total} & 616 & 329289 & 323672 & 3635 & 1672 & 658268 & 25.7\% \\
        
        \bottomrule
    \end{tabular}
    \caption{\textbf{Summary of annotated videos and samples from each surgical procedure and task.} We outline the number of videos and samples in each task for each procedure and specialty. We also present the distribution of manual annotations at the procedure level. For the exact surgical component tags within each task, please refer to the Halsted Surgical Atlas section.}
    \label{table:data}
\end{table*}

\begin{table*}[!t]
\centering
\renewcommand{\arraystretch}{1.2}
\begin{tabular}{llp{10cm}c}
\hline
\textbf{Task} & \textbf{Tag} & \textbf{Categories} & \textbf{\# Categories} \\
\hline
\multirow{7}{*}{1}
 & Specialty & Bariatric, Cardiac, Colorectal, General, Gynecology, Hepatobiliary, Thoracic, Urology & 8 \\
 & Procedure / Method & Gastric Bypass, Atrial Closure, IMA Harvest, Mitral Stitching, Right Colectomy, Total Mesorectal Excision, Laparoscopic Cholecystectomy, Laparoscopic Hernia, Endometriosis, Hysterectomy, Myomectomy, Pancreaticoduodenectomy, Right Middle Lobectomy, Right Upper Lobectomy, Segmentectomy, Prostatectomy & 16 \\
 & Step & Suturing, Dissection & 2 \\
\hline
\multirow{11}{*}{2} 
 & Action & Assistant, Cold Cut, Cautery, Extraction, Fluorescence, Hot Cut, Hook, Idle, Clip, Camera Move, Mesh, Push/Peel, Retraction, Spread, Sponge, Stapler, Tube, Tug, Other & 19 \\ & Arm & Left, Right, Both & 3 \\
 & Instrument & Bipolar Dissector, Bipolar Forceps, Bipolar Forceps–Cautery Hook, Bipolar Forceps–Monopolar Scissors, Bipolar Forceps–Vessel Sealer, Bipolar Grasper, Bipolar Grasper–Monopolar Scissors, Cadiere Forceps, Cadiere Forceps–Bipolar Grasper, Cautery Spatula, Cautery Hook, Clip Applier, Clipper, Fenestrated Forceps, Fenestrated Grasper, Fenestrated Grasper–Bipolar Grasper, Grasper, Hook Monopolar, Maryland Grasper, Monopolar Scissors, Needle Driver, Prograsp Forceps, Scissors, Shears, Stapler, Suction, Vessel Sealer & 27 \\
\hline
\multirow{2}{*}{3} 
 & Phase & Needle Handling, Driving, Withdrawal & 3 \\
 & Proficiency & Low, High & 2 \\
\hline
\multirow{6}{*}{4} 
 & Anatomy & Bile Duct, Bile Duct–Small Intestine, Bladder, Bladder–Urethra, Colon, Left Atrium, Mitral Annulus, Pancreas, Pancreas–Small Intestine, Peritoneum, Small Intestine, Small Intestine–Bile Duct, Small Intestine–Stomach, Stomach, Stomach–Small Intestine, Urethra, Uterus, Vagina & 18 \\
 & Extent of Stitch & Single, Double, Surface & 3 \\
 & Directionality & In, Out, Both & 3 \\
\hline
\end{tabular}
\caption{\textbf{Breakdown of each task, component tags, and categories.} The tags span four tasks: macro-activity, micro-activity, proficiency, and context. Each tag comprises a set of discrete categories. Hyphenated categories indicate items that co-occur.}
\label{table:task_breakdown}
\end{table*}

\subsection*{Halsted surgical atlas}

We trained Halsted on the Halsted Surgical Atlas (HSA), the largest library of its kind, comprising 650K+ video clips and corresponding surgical component tag annotations (see Table~\ref{table:data}). We sourced the videos from the public domain (predominantly through YouTube) which span 16 procedures across 8 surgical specialties (see Table~\ref{table:data}). To annotate these videos with surgical component tags, we leveraged a trained annotator with over 4 years of experience annotating surgical videos according to peer-reviewed and established visual taxonomies previously described \cite{Kiyasseh2023, Haque2022}. These taxonomies present a discrete set of surgical actions and visual criteria needed to assess suturing proficiency. In light of the accessibility of these annotation rubrics, a trained annotator was used to manually annotate a subset of the surgical video clips (see Table~\ref{table:data} for a breakdown of manual annotations). Although no rubric is perfect, we opted for these due to the precedent set in previous publications \cite{Kiyasseh2023, Kiyasseh2023CommsMed, Kiyasseh2023DigMed}, their recently-demonstrated relationship with post-operative patient outcomes \cite{Heard2025}, and the ease with which they can be followed by annotators. It is worthwhile to note that Halsted is also amenable to working with any other taxonomy, as we demonstrate in the external validation section. 

\begin{figure*}[!h]
    \centering
    \includegraphics[width=1\linewidth]{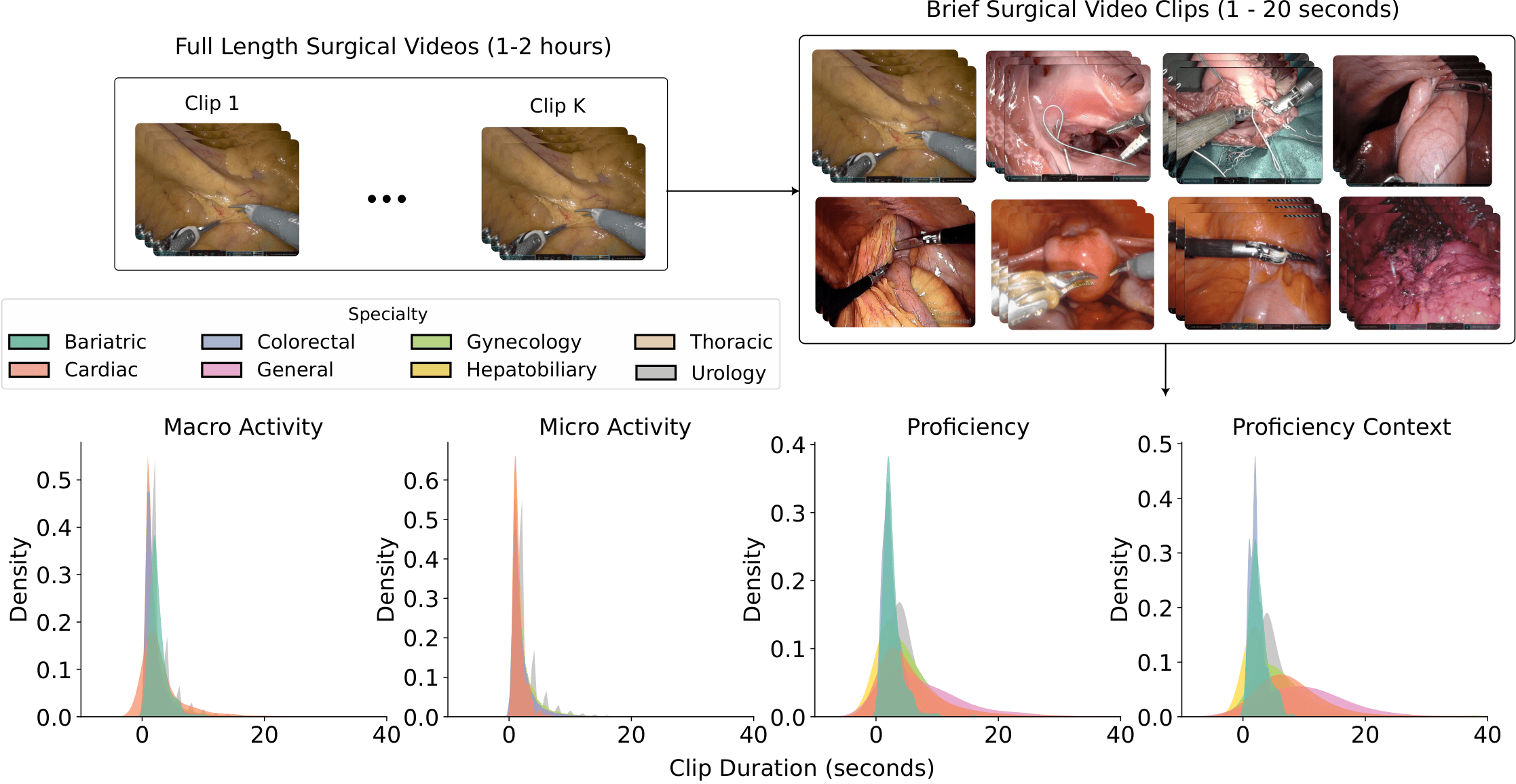}
    \caption{\textbf{Distribution of durations of video clips in the Halsted Surgical Atlas.} We collect full-length surgical videos (1-2 hours in duration) which comprise a sequence of video clips and curate video clips based on task-specific timestamps and annotations (1-20 seconds in duration). We show the distribution of the duration of these video clips across the four tasks and 8 surgical specialties. During training and evaluation of the Halsted model, we ensure the full-length surgical videos only appear in one of the training, validation, or test sets to avoid video contamination of the results.}
    \label{fig:data_stats}
\end{figure*}

We manually annotated a subset of the surgical videos in the Halsted surgical atlas, forming curated surgical video clips from full-length surgical videos (see Fig.~\ref{fig:data_stats}). Depending on the taxonomy used for annotation (e.g., micro-activity), we assigned the surgical video clips to distinct tasks (see Fig.~\ref{fig:halsted}). Each task is associated with a sequence of component tags and each tag comprises a set of categories (Table~\ref{table:task_breakdown}). For example, the micro-activity task is associated with the component tags of action, arm, and instrument where the action component comprises 19 discrete categories, etc. For a full breakdown of the tasks, component tags, and categories, please refer to Table~\ref{table:task_breakdown}.

Depending on the visual content of a video clip, it might only be associated with a subset of surgical component tags. For example, a video clip of a surgeon dissecting tissue will not have any suturing-related tags. Nonetheless, the variability in videos across surgical specialties and the wide range of surgical component tags provide a comprehensive basis for the training and evaluation of Halsted.

\begin{figure*}[!t]
    \centering
    \includegraphics[width=1\linewidth]{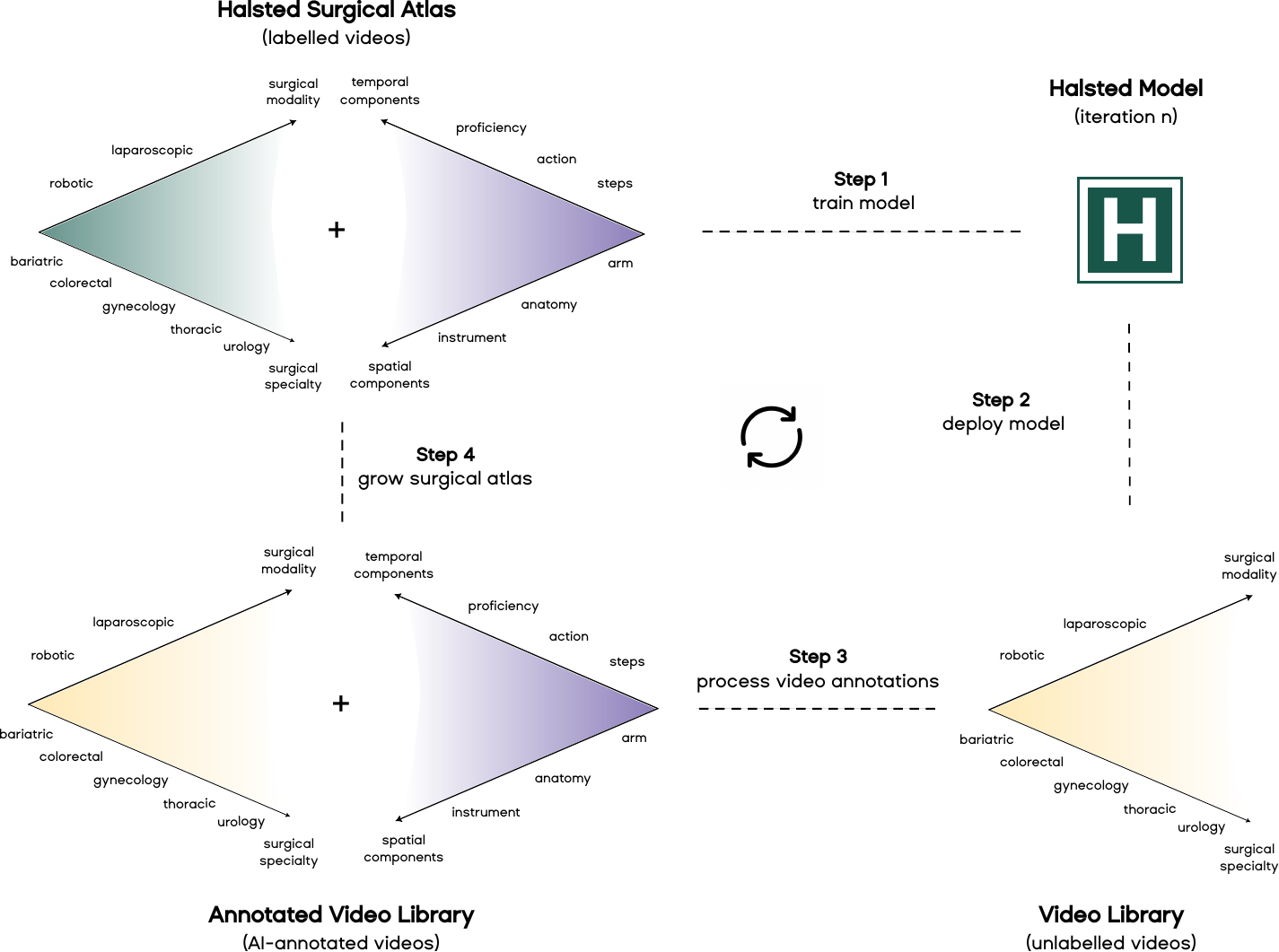}
    \caption{\textbf{Self-labelling strategy to curate the Halsted Surgical Atlas.} We curated the atlas iteratively, repeating a sequence of four steps. We started with manually-curated video library annotated with component tags, trained a model on the video library, and deployed it on unannotated surgical videos to obtain AI-annotations. We then added the newly-processed videos to the original video library in preparation for training the model on a larger volume dataset. The cycle can repeat with the collection of new surgical videos.}
    \label{fig:halsted_training_stages}
\end{figure*}

\textbf{Iterative development of the Halsted surgical atlas.} In pursuit of a high-quality and high-volume video library, we curated the Halsted Surgical Atlas iteratively, repeating a sequence of four steps (see Fig.~\ref{fig:halsted_training_stages}). In the first cycle, we started with a video library comprising video clips from a pre-defined set of surgical specialities and procedures that were manually-annotated with surgical component tags. We opted to annotate a subset of video clips across all procedures instead of annotating all video clips associated with a single procedure. The motivation was to maximize the data exposure of our model, ensuring it can reliably map the entire space of procedures. 

\textbf{Step 1.} We trained the first version of Halsted on this carefully-curated manually-annotated multi-specialty video library, and measured its performance in generating the surgical component tags. When Halsted achieved an average accuracy of 80\% across all tasks, we gained sufficient confidence in its ability to generate annotations for unseen videos. 

\textbf{Step 2.} We then deployed Halsted on the remaining videos without annotations. Specifically, we split full-length videos into 1-second non-overlapping segments to create 1-second video clips. Halsted was presented with these clips alongside task instructions to generate surgical component tags (i.e., AI-generated annotations). We focused on the micro-activity task (see Fig.~\ref{fig:halsted_performance}) given its temporally fine-grained annotations and the need for substantial samples to achieve strong performance. At this point, we had video clips newly-annotated with micro-activity. 

\textbf{Step 3.} To ensure high-quality annotations, we post-processed the newly-annotated video clips. First, we removed predictions with low confidence in the action taken during that 1-second window. Each action had its own confidence threshold which we determined based on performance on the validation set. Second, and after removing low-confidence predictions, we set out to define the temporal boundary of the remaining predictions. We merged separate AI-generated annotations if they reflected the same action within a 1-second time-window. This ensured we avoided mini-actions that were visually incomplete and enabled us to capture the full duration of an action. 

\textbf{Step 4.} We added this quality-controlled AI-annotated video library to the original manually-curated video library, growing it in size and forming the Halsted Surgical Atlas. The video library grew from 168,912 video samples to 658,268 video samples. We used the expanded video library to train the second version of Halsted. Although such a cycle can be repeated with each additional collection of surgical videos, we conducted it once, increasing the size of the dataset four-fold in the process.

\subsection*{Evaluating the model}

\textbf{Monte-Carlo cross-validation.} We evaluate Halsted using 5-fold Monte-Carlo cross-validation with a leave-one-video-out evaluation setup. In each fold, we selected two unique videos and placed one in the validation set and the other in the test set. All remaining videos, and their corresponding video samples, were used for training the model. By ensuring videos do not appear in more than one set, we avoid video contamination and reduce the likelihood of over-estimating performance. For an exact number of video samples in each fold, task, and set, we refer readers to Supplementary Note 1. Unless explicitly stated otherwise, we always report performance on the test set of the five folds.

\textbf{HSA-27k.} To facilitate benchmarking, we open-source a subset of the Halsted Surgical Atlas (HSA-27k) which comprises approximately 27k video clips. HSA-27k is the test set of the first fold used in the aforementioned 5-fold cross-validation setup. An exact breakdown of the number of video samples in each task and category can be found in Supplementary Note 2. The dataset can be accessed on HuggingFace: \url{https://huggingface.co/datasets/halsted-ai/halsted-surgical-atlas}.

\textbf{Evaluation metrics.} We designed the model to comprehensively map multiple components of surgery. To evaluate model performance, we chose two types of metrics; those that quantify the model's ability to distinguish between distinct categories given video samples with pre-defined temporal boundaries (accuracy and area under the receiver operating characteristic curve) and others that account for temporal boundaries (temporal F1 score). We measure accuracy for each surgical component by quantifying the proportion of ground-truth component tags that perfectly match the tags generated by Halsted. To measure the temporal F1 score for a surgical component, we first quantify the level of temporal overlap between the ground-truth time-window of that surgical component tag and the time-window generated by Halsted. Akin to previous work \cite{Psychogyios2023, Kiyasseh2023}, we treat a 10\% overlap or more in those time-windows as a true positive (i.e., a match). Anything less than that is considered either a false positive, if a prediction is made when no ground-truth time-window exists, or a false negative, if a prediction is not made when a ground-truth time-window exists. 

It is worthwhile to note that metrics like accuracy, where time-windows are not factored into the evaluation process, are a critical first step in evaluation since a model that is unable to distinguish between pre-defined video samples will surely flounder when tasked with factoring in temporal boundaries. Specifically, a weak AUROC portends a weak temporal F1-score. Although no metric is perfect \cite{Reinke2025}, and measuring accuracy can be flawed, we chose it precisely because of the greedy decoding mechanism we adopt during inference whereby we select the most likely token at each time step during the autoregressive rollout.

\subsection*{Performing inference with the model}

\textbf{Controllable generation of surgical component tags} To control the type of component tags generated by our model (multitask behaviour), we can simply replace the input instructions $I$. We assign each task $k \in \{1, ..., K \}$ to a unique instruction $I_{k}$ and, during training, expose the model to all such instructions enabling it to perform \textit{all} tasks (see Fig.~\ref{fig:halsted}, left). During inference, however, we can finely control the model's outputs by providing it with one specific instruction. This controllability is desirable as it ensures model outputs are aligned with a user's intentions yet it remains a missing component of existing models.

\textbf{Mapping the surgical component tags.} At each time step in the sequential generation process, we obtain a probability distribution over the vocabulary of surgical component tokens $p_{V_{A}} \in R^{V_{A}}$. To sample from this distribution, we adopt greedy decoding, selecting the most likely token at each time step. We retrieve the sampled token's embedding before feeding it into the backbone for the subsequent time step.

\subsection*{Benchmarking against state-of-the-art model}

We compared Halsted's ability to map surgery to that of SAIS \cite{Kiyasseh2023}. To enable a fair comparison, we trained both models only on RGB videos, omitting the optical flow input outlined in the original model. After discovering that SAIS struggled to distinguish between the the action categories (see Table~\ref{table:task_breakdown}, Task 2), we opted to focus on a task with fewer categories, namely recognizing suturing steps and assessing suturing performance. Even for such tasks, we found that SAIS struggled when videos from all eight specialties were presented simultaneously. We therefore chose three specialties at random (bariatric, general, gynecology) to enable SAIS to learn something at the very least. We split this subset into training, validation, and test sets ensuring that videos do not appear in more than one set and report metrics on the test set (see Implementation details in Methods). Halsted and SAIS were then trained on the same exact training set and evaluated on the held-out set, ensuring a fair performance comparison.

\subsection*{Investigating the effect of decoder size}

To investigate the effect of the decoder size on Halsted's performance, we performed a drop-in replacement of the 2-layer transformer decoder (44M parameters) with a pre-trained LLama-3.2 decoder (1B parameters). We used the same vision encoder to extract representations from video clips, and akin to other multimodal models \cite{Lin2024}, we include an adaptor module in the form of a multi-layer perceptron to project the video representation into the embedding space of the language tokens. We fine-tune the adaptor module and language model parameters and adopt the same training and evaluation setup, with Monte-Carlo cross-validation, ensuring reporting performance on the held-out test sets. The total number of trainable parameters therefore amounts to 1.2B parameters. Halsted with Llama is trained to solve the micro-activity task, generating a sequence of surgical action, arm, and instrument for each video clip as with the standard Halsted model. Instead of our custom word-level tokenizer, we use the standard Llama byte-level byte-pair encoding tokenizer with approximately 128k tokens. To evaluate the model and measure its accuracy, we generate a full sequence of surgical component tags for each video clip, use a regular expression to extract the generated component tag (e.g., <action>), and compare it to the ground-truth tag.

\subsection*{Visualizing Halsted representations}

We tasked Halsted with assessing suturing proficiency on the held-out test set of surgical video clips. We extracted the representations of all inputs at the final layer of the transformer decoder and averaged them across the sequence dimension, resulting in a single representation for each video clip. We apply UMAP to these representations (Halsted Features) and obtain the two-dimensional embeddings which are displayed in Fig.~\ref{fig:halsted_embeddings} (right). Using the same held-out set of surgical video clips, we separately apply UMAP to the features extracted from VideoMAE (VideoMAE Features) and which are displayed in Fig.~\ref{fig:halsted_embeddings} (left).

\subsection*{External validation of Halsted}

We benchmarked the Halsted model on the publicly-available RARP-50 dataset \cite{Psychogyios2023} which comprises surgical videos of the dorsal venous complex step of a prostatectomy and corresponding annotations of the group of eight actions taken to perform that step. As Halsted is a video-based model, we focus on the provided video-level annotations (and not the frame-level annotations) which outline the start and end frames of the actions. RARP-50 is split into a training set of 40 videos and a test set of 10 videos. We train Halsted on the training set and report performance on the held-out test set. To facilitate comparison with previous methods \cite{Van2020, Kiyasseh2023}, we report the average accuracy of the model on the test-set across all videos and action categories. We follow the same training process outlined in the implementation details, except that we solve for a single task with a single component (action recognition). We experiment with both $L=2$ and $L=10$ layer transformer decoder and obtain an accuracy of $66.6\%$ and $68.6\%$ on the test set, respectively. We found that a deeper decoder does not confer additional performance benefits.

We compare Halsted's performance to those reported for competing methods including MA-TCN \cite{Van2022} and SAIS \cite{Kiyasseh2023}. Since the latter adopt a slightly different evaluation setup, notably filtering out infrequent actions from the dataset, we also report the relative improvement in accuracy of these models compared to a random chance classifier. We believe doing so better contextualizes the performance of the competing methods as it accounts for the number of surgical action categories the model is expected to identify. Further, the accuracy reported in the associated challenge paper \cite{Psychogyios2023} is based on frame-level action predictions and is therefore not comparable to our reported metrics based on video-level predictions. Our reported accuracy metric can therefore act as a baseline for future researchers who develop and evaluate video-based models.

\subsection*{Halsted platform}

We developed the Halsted platform, a web application powered by the Halsted model, to enable surgeons to map their own procedures from surgical videos. Surgeons can upload a video of any duration (e.g., on the order of hours) and select the type of automated mapping they are most interested in, coinciding with the tasks outlined in Fig.~\ref{fig:halsted}. The Halsted model, deployed in cloud compute servers, automatically maps the surgery and provides surgeons with a personalized dashboard displaying surgical components including the actions they have taken, robotic arms used, instruments deployed, and the proficiency with which they performed such actions.

\textbf{Multi-stage workflow for Halsted model.} To account for the variability and extended duration of full-length surgical videos, which can be on order of hours, and to ensure the Halsted model generates high-quality outputs, we adopted a multi-stage workflow with a sequence of pre- and post-processing steps. We use the logic that Halsted will need to process a larger temporal window in a video to recognize coarse surgical components (e.g., step of a procedure) than granular components (e.g., actions and instruments used). When a surgeon selects the type of mapping they are interested in, the platform initiates a multi-stage workflow where Halsted is first presented with 30-second non-overlapping windows of the uploaded video and is tasked with identifying coarse components (e.g., suturing). Since this output is also timestamped, we can pinpoint the exact temporal occurrence of the surgical components. If a surgeon has chosen to map suturing components, then we only focus on the temporal segments of the video which Halsted has identified as depicting suturing activity. Our multi-stage workflow therefore comprises a built-in quality assurance mechanism to ensure subsequent processing steps are only applied to the most relevant temporal segments of the video. This approach confers additional benefits beyond quality assurance, (1) reducing the time-taken to process the video and ensuring surgeons receive a quick result and (2) reducing the inference cost compared to naively processing the entire video. Halsted is now presented with fine-grained 2-5 second non-overlapping windows of the segmented video alongside a new set of task instructions to map granular surgical components (e.g., anatomy and surgical proficiency). In this multi-stage workflow, we iteratively expand the comprehensiveness of the surgical mapping while ensuring high-quality predictions at each step of the way.

\subsection*{Reporting summary} 
Further information on research design is available in the Nature Research Reporting Summary linked to this article.

\subsection*{Data availability}
To facilitate benchmarking, we have made a subset of the Halsted Surgical Atlas, HSA-27K, available to the public. It can be accessed on HuggingFace (\url{https://huggingface.co/datasets/halsted-ai/halsted-surgical-atlas}).

\subsection*{Code availability}
We make the Halsted model accessible via the Halsted platform (\url{https://halstedhealth.ai/}) and a custom Python SDK (\url{https://docs.halstedhealth.ai/}).



\subsection*{Author contributions}
D.K. conceived of and designed the study, curated the data, developed the model and platform, evaluated the results, and wrote the manuscript. 

\subsection*{Competing Interests}
Halsted AI has filed for patent protection for D.K. for the work related to the model and platform.


\begin{thebibliography}{10}
\urlstyle{rm}
\expandafter\ifx\csname url\endcsname\relax
  \def\url#1{\texttt{#1}}\fi
\expandafter\ifx\csname urlprefix\endcsname\relax\def\urlprefix{URL }\fi
\expandafter\ifx\csname doiprefix\endcsname\relax\def\doiprefix{DOI: }\fi
\providecommand{\bibinfo}[2]{#2}
\providecommand{\eprint}[2][]{\url{#2}}

\bibitem{Birkmeyer2013}
\bibinfo{author}{Birkmeyer, J.~D.} \emph{et~al.}
\newblock \bibinfo{journal}{\bibinfo{title}{Surgical skill and complication rates after bariatric surgery}}.
\newblock {\emph{\JournalTitle{New England Journal of Medicine}}} \textbf{\bibinfo{volume}{369}}, \bibinfo{pages}{1434--1442} (\bibinfo{year}{2013}).

\bibitem{Stulberg2020}
\bibinfo{author}{Stulberg, J.~J.} \emph{et~al.}
\newblock \bibinfo{journal}{\bibinfo{title}{Association between surgeon technical skills and patient outcomes}}.
\newblock {\emph{\JournalTitle{JAMA Surgery}}} \textbf{\bibinfo{volume}{155}}, \bibinfo{pages}{960--968} (\bibinfo{year}{2020}).

\bibitem{Volpe2015}
\bibinfo{author}{Volpe, A.} \emph{et~al.}
\newblock \bibinfo{journal}{\bibinfo{title}{Pilot validation study of the european association of urology robotic training curriculum}}.
\newblock {\emph{\JournalTitle{European Urology}}} \textbf{\bibinfo{volume}{68}}, \bibinfo{pages}{292--299} (\bibinfo{year}{2015}).

\bibitem{Valdis2016}
\bibinfo{author}{Valdis, M.}, \bibinfo{author}{Chu, M.~W.}, \bibinfo{author}{Schlachta, C.} \& \bibinfo{author}{Kiaii, B.}
\newblock \bibinfo{journal}{\bibinfo{title}{Evaluation of robotic cardiac surgery simulation training: a randomized controlled trial}}.
\newblock {\emph{\JournalTitle{The Journal of Thoracic and Cardiovascular Surgery}}} \textbf{\bibinfo{volume}{151}}, \bibinfo{pages}{1498--1505} (\bibinfo{year}{2016}).

\bibitem{Kiely2015}
\bibinfo{author}{Kiely, D.~J.} \emph{et~al.}
\newblock \bibinfo{journal}{\bibinfo{title}{Virtual reality robotic surgery simulation curriculum to teach robotic suturing: a randomized controlled trial}}.
\newblock {\emph{\JournalTitle{Journal of Robotic Surgery}}} \textbf{\bibinfo{volume}{9}}, \bibinfo{pages}{179--186} (\bibinfo{year}{2015}).

\bibitem{Sobel2016}
\bibinfo{author}{Sobel, R.~H.} \emph{et~al.}
\newblock \bibinfo{journal}{\bibinfo{title}{Implementation of a comprehensive competency-based transoral robotic surgery training curriculum with ex vivo dissection models}}.
\newblock {\emph{\JournalTitle{Head \& Neck}}} \textbf{\bibinfo{volume}{38}}, \bibinfo{pages}{1553--1563} (\bibinfo{year}{2016}).

\bibitem{Childers2023}
\bibinfo{author}{Childers, C.~P.} \emph{et~al.}
\newblock \bibinfo{journal}{\bibinfo{title}{Longitudinal trends in efficiency and complexity of surgical procedures: Analysis of 1.7 million operations between 2019 and 2023}}.
\newblock {\emph{\JournalTitle{Journal of the American College of Surgeons}}} \bibinfo{pages}{10--1097} (\bibinfo{year}{2023}).

\bibitem{Zia2018}
\bibinfo{author}{Zia, A.}, \bibinfo{author}{Sharma, Y.}, \bibinfo{author}{Bettadapura, V.}, \bibinfo{author}{Sarin, E.~L.} \& \bibinfo{author}{Essa, I.}
\newblock \bibinfo{journal}{\bibinfo{title}{Video and accelerometer-based motion analysis for automated surgical skills assessment}}.
\newblock {\emph{\JournalTitle{International Journal of Computer Assisted Radiology and Surgery}}} \textbf{\bibinfo{volume}{13}}, \bibinfo{pages}{443--455} (\bibinfo{year}{2018}).

\bibitem{Funke2019}
\bibinfo{author}{Funke, I.} \emph{et~al.}
\newblock \bibinfo{title}{Using 3d convolutional neural networks to learn spatiotemporal features for automatic surgical gesture recognition in video}.
\newblock In \emph{\bibinfo{booktitle}{International Conference on Medical Image Computing and Computer-Assisted Intervention}}, \bibinfo{pages}{467--475} (\bibinfo{organization}{Springer}, \bibinfo{year}{2019}).

\bibitem{Kiyasseh2023}
\bibinfo{author}{Kiyasseh, D.} \emph{et~al.}
\newblock \bibinfo{journal}{\bibinfo{title}{A vision transformer for decoding surgeon activity from surgical videos}}.
\newblock {\emph{\JournalTitle{Nature Biomedical Engineering}}} \textbf{\bibinfo{volume}{7}}, \bibinfo{pages}{780--796} (\bibinfo{year}{2023}).

\bibitem{Kiyasseh2023CommsMed}
\bibinfo{author}{Kiyasseh, D.} \emph{et~al.}
\newblock \bibinfo{journal}{\bibinfo{title}{A multi-institutional study using artificial intelligence to provide reliable and fair feedback to surgeons}}.
\newblock {\emph{\JournalTitle{Communications Medicine}}} \textbf{\bibinfo{volume}{3}}, \bibinfo{pages}{42} (\bibinfo{year}{2023}).

\bibitem{Kiyasseh2023DigMed}
\bibinfo{author}{Kiyasseh, D.} \emph{et~al.}
\newblock \bibinfo{journal}{\bibinfo{title}{Human visual explanations mitigate bias in ai-based assessment of surgeon skills}}.
\newblock {\emph{\JournalTitle{NPJ Digital Medicine}}} \textbf{\bibinfo{volume}{6}}, \bibinfo{pages}{54} (\bibinfo{year}{2023}).

\bibitem{Yuan2025}
\bibinfo{author}{Yuan, K.} \emph{et~al.}
\newblock \bibinfo{journal}{\bibinfo{title}{Learning multi-modal representations by watching hundreds of surgical video lectures}}.
\newblock {\emph{\JournalTitle{Medical Image Analysis}}} \bibinfo{pages}{103644} (\bibinfo{year}{2025}).

\bibitem{Hu2025}
\bibinfo{author}{Hu, M.} \emph{et~al.}
\newblock \bibinfo{title}{Ophclip: Hierarchical retrieval-augmented learning for ophthalmic surgical video-language pretraining}.
\newblock In \emph{\bibinfo{booktitle}{Proceedings of the IEEE/CVF International Conference on Computer Vision}}, \bibinfo{pages}{19838--19849} (\bibinfo{year}{2025}).

\bibitem{Meara2015}
\bibinfo{author}{Meara, J.~G.} \emph{et~al.}
\newblock \bibinfo{journal}{\bibinfo{title}{Global surgery 2030: evidence and solutions for achieving health, welfare, and economic development}}.
\newblock {\emph{\JournalTitle{The Lancet}}} \textbf{\bibinfo{volume}{386}}, \bibinfo{pages}{569--624} (\bibinfo{year}{2015}).

\bibitem{Reinke2025}
\bibinfo{author}{Reinke, A.} \emph{et~al.}
\newblock \bibinfo{journal}{\bibinfo{title}{Current validation practice undermines surgical ai development}}.
\newblock {\emph{\JournalTitle{arXiv preprint arXiv:2511.03769}}}  (\bibinfo{year}{2025}).

\bibitem{Haque2022}
\bibinfo{author}{Haque, T.~F.} \emph{et~al.}
\newblock \bibinfo{journal}{\bibinfo{title}{An assessment tool to provide targeted feedback to robotic surgical trainees: Development and validation of the end-to-end assessment of suturing expertise (ease)}}.
\newblock {\emph{\JournalTitle{Urology Practice}}} \bibinfo{pages}{10--1097} (\bibinfo{year}{2022}).

\bibitem{Tong2022}
\bibinfo{author}{Tong, Z.}, \bibinfo{author}{Song, Y.}, \bibinfo{author}{Wang, J.} \& \bibinfo{author}{Wang, L.}
\newblock \bibinfo{journal}{\bibinfo{title}{Videomae: Masked autoencoders are data-efficient learners for self-supervised video pre-training}}.
\newblock {\emph{\JournalTitle{Advances in Neural Information Processing Systems}}} \textbf{\bibinfo{volume}{35}}, \bibinfo{pages}{10078--10093} (\bibinfo{year}{2022}).

\bibitem{Mcinnes2018}
\bibinfo{author}{McInnes, L.}, \bibinfo{author}{Healy, J.} \& \bibinfo{author}{Melville, J.}
\newblock \bibinfo{journal}{\bibinfo{title}{Umap: Uniform manifold approximation and projection for dimension reduction}}.
\newblock {\emph{\JournalTitle{arXiv preprint arXiv:1802.03426}}}  (\bibinfo{year}{2018}).

\bibitem{Van2022}
\bibinfo{author}{Van~Amsterdam, B.} \emph{et~al.}
\newblock \bibinfo{journal}{\bibinfo{title}{Gesture recognition in robotic surgery with multimodal attention}}.
\newblock {\emph{\JournalTitle{IEEE Transactions on Medical Imaging}}}  (\bibinfo{year}{2022}).

\bibitem{Heard2025}
\bibinfo{author}{Heard, J.~R.} \emph{et~al.}
\newblock \bibinfo{journal}{\bibinfo{title}{Surgical performance metrics for 1-year patient-reported outcomes after radical prostatectomy}}.
\newblock {\emph{\JournalTitle{JAMA Surgery}}} \textbf{\bibinfo{volume}{160}}, \bibinfo{pages}{674--680} (\bibinfo{year}{2025}).

\bibitem{Ye2025}
\bibinfo{author}{Ye, Z.} \emph{et~al.}
\newblock \bibinfo{journal}{\bibinfo{title}{A comprehensive video dataset for surgical laparoscopic action analysis}}.
\newblock {\emph{\JournalTitle{Scientific Data}}} \textbf{\bibinfo{volume}{12}}, \bibinfo{pages}{862} (\bibinfo{year}{2025}).

\bibitem{Derathe2025}
\bibinfo{author}{Derath{\'e}, A.} \emph{et~al.}
\newblock \bibinfo{journal}{\bibinfo{title}{Lapex: A new multimodal dataset for context recognition and practice assessment in laparoscopic surgery}}.
\newblock {\emph{\JournalTitle{Scientific Data}}} \textbf{\bibinfo{volume}{12}}, \bibinfo{pages}{342} (\bibinfo{year}{2025}).

\bibitem{Carstens2023}
\bibinfo{author}{Carstens, M.} \emph{et~al.}
\newblock \bibinfo{journal}{\bibinfo{title}{The dresden surgical anatomy dataset for abdominal organ segmentation in surgical data science}}.
\newblock {\emph{\JournalTitle{Scientific Data}}} \textbf{\bibinfo{volume}{10}}, \bibinfo{pages}{1--8} (\bibinfo{year}{2023}).

\bibitem{Nwoye2022activity}
\bibinfo{author}{Nwoye, C.~I.} \emph{et~al.}
\newblock \bibinfo{journal}{\bibinfo{title}{Cholec{T}riplet2021: A benchmark challenge for surgical action triplet recognition}}.
\newblock {\emph{\JournalTitle{Preprint at https://arxiv.org/abs/2204.04746}}}  (\bibinfo{year}{2022}).

\bibitem{Goodman2024}
\bibinfo{author}{Goodman, E.~D.} \emph{et~al.}
\newblock \bibinfo{journal}{\bibinfo{title}{Analyzing surgical technique in diverse open surgical videos with multitask machine learning}}.
\newblock {\emph{\JournalTitle{JAMA Surgery}}} \textbf{\bibinfo{volume}{159}}, \bibinfo{pages}{185--192} (\bibinfo{year}{2024}).

\bibitem{Ghamsarian2024}
\bibinfo{author}{Ghamsarian, N.} \emph{et~al.}
\newblock \bibinfo{journal}{\bibinfo{title}{Cataract-1k dataset for deep-learning-assisted analysis of cataract surgery videos}}.
\newblock {\emph{\JournalTitle{Scientific Data}}} \textbf{\bibinfo{volume}{11}}, \bibinfo{pages}{373} (\bibinfo{year}{2024}).

\bibitem{Wang2023}
\bibinfo{author}{Wang, W.} \emph{et~al.}
\newblock \bibinfo{journal}{\bibinfo{title}{Visionllm: Large language model is also an open-ended decoder for vision-centric tasks}}.
\newblock {\emph{\JournalTitle{Advances in Neural Information Processing Systems}}} \textbf{\bibinfo{volume}{36}}, \bibinfo{pages}{61501--61513} (\bibinfo{year}{2023}).

\bibitem{Song2024}
\bibinfo{author}{Song, E.} \emph{et~al.}
\newblock \bibinfo{title}{Moviechat: From dense token to sparse memory for long video understanding}.
\newblock In \emph{\bibinfo{booktitle}{Proceedings of the IEEE/CVF Conference on Computer Vision and Pattern Recognition}}, \bibinfo{pages}{18221--18232} (\bibinfo{year}{2024}).

\bibitem{Yang2023}
\bibinfo{author}{Yang, A.} \emph{et~al.}
\newblock \bibinfo{title}{Vid2seq: Large-scale pretraining of a visual language model for dense video captioning}.
\newblock In \emph{\bibinfo{booktitle}{Proceedings of the IEEE/CVF Conference on Computer Vision and Pattern Recognition}}, \bibinfo{pages}{10714--10726} (\bibinfo{year}{2023}).

\bibitem{Liu2024}
\bibinfo{author}{Liu, H.}, \bibinfo{author}{Li, C.}, \bibinfo{author}{Li, Y.} \& \bibinfo{author}{Lee, Y.~J.}
\newblock \bibinfo{title}{Improved baselines with visual instruction tuning}.
\newblock In \emph{\bibinfo{booktitle}{Proceedings of the IEEE/CVF Conference on Computer Vision and Pattern Recognition}}, \bibinfo{pages}{26296--26306} (\bibinfo{year}{2024}).

\bibitem{Lu2024}
\bibinfo{author}{Lu, J.} \emph{et~al.}
\newblock \bibinfo{title}{Unified-io 2: Scaling autoregressive multimodal models with vision language audio and action}.
\newblock In \emph{\bibinfo{booktitle}{Proceedings of the IEEE/CVF Conference on Computer Vision and Pattern Recognition}}, \bibinfo{pages}{26439--26455} (\bibinfo{year}{2024}).

\bibitem{Chen2022}
\bibinfo{author}{Chen, T.}, \bibinfo{author}{Saxena, S.}, \bibinfo{author}{Li, L.}, \bibinfo{author}{Fleet, D.~J.} \& \bibinfo{author}{Hinton, G.}
\newblock \bibinfo{title}{Pix2seq: A language modeling framework for object detection}.
\newblock In \emph{\bibinfo{booktitle}{International Conference on Learning Representations}} (\bibinfo{year}{2022}).

\bibitem{Wei2025}
\bibinfo{author}{Wei, J.} \emph{et~al.}
\newblock \bibinfo{journal}{\bibinfo{title}{Surgbench: A unified large-scale benchmark for surgical video analysis}}.
\newblock {\emph{\JournalTitle{arXiv preprint arXiv:2506.07603}}}  (\bibinfo{year}{2025}).

\bibitem{Schmidgall2024}
\bibinfo{author}{Schmidgall, S.}, \bibinfo{author}{Kim, J.~W.}, \bibinfo{author}{Jopling, J.} \& \bibinfo{author}{Krieger, A.}
\newblock \bibinfo{journal}{\bibinfo{title}{General surgery vision transformer: A video pre-trained foundation model for general surgery}}.
\newblock {\emph{\JournalTitle{arXiv preprint arXiv:2403.05949}}}  (\bibinfo{year}{2024}).

\bibitem{Yang2025}
\bibinfo{author}{Yang, S.} \emph{et~al.}
\newblock \bibinfo{journal}{\bibinfo{title}{Large-scale self-supervised video foundation model for intelligent surgery}}.
\newblock {\emph{\JournalTitle{arXiv preprint arXiv:2506.02692}}}  (\bibinfo{year}{2025}).

\bibitem{Jaspers2025}
\bibinfo{author}{Jaspers, T.~J.} \emph{et~al.}
\newblock \bibinfo{journal}{\bibinfo{title}{Scaling up self-supervised learning for improved surgical foundation models}}.
\newblock {\emph{\JournalTitle{arXiv preprint arXiv:2501.09436}}}  (\bibinfo{year}{2025}).

\bibitem{Che2025}
\bibinfo{author}{Che, C.}, \bibinfo{author}{Wang, C.}, \bibinfo{author}{Vercauteren, T.}, \bibinfo{author}{Tsoka, S.} \& \bibinfo{author}{Garcia-Peraza-Herrera, L.~C.}
\newblock \bibinfo{journal}{\bibinfo{title}{Surg-3m: A dataset and foundation model for perception in surgical settings}}.
\newblock {\emph{\JournalTitle{arXiv preprint arXiv:2503.19740}}}  (\bibinfo{year}{2025}).

\bibitem{Fazlollahi2022}
\bibinfo{author}{Fazlollahi, A.~M.} \emph{et~al.}
\newblock \bibinfo{journal}{\bibinfo{title}{Effect of artificial intelligence tutoring vs expert instruction on learning simulated surgical skills among medical students: a randomized clinical trial}}.
\newblock {\emph{\JournalTitle{JAMA Network Open}}} \textbf{\bibinfo{volume}{5}}, \bibinfo{pages}{e2149008--e2149008} (\bibinfo{year}{2022}).

\bibitem{Giglio2025}
\bibinfo{author}{Giglio, B.} \emph{et~al.}
\newblock \bibinfo{journal}{\bibinfo{title}{Artificial intelligence--augmented human instruction and surgical simulation performance: a randomized clinical trial}}.
\newblock {\emph{\JournalTitle{JAMA Surgery}}} \textbf{\bibinfo{volume}{160}}, \bibinfo{pages}{993--1003} (\bibinfo{year}{2025}).

\bibitem{Schlick2020}
\bibinfo{author}{Schlick, C. J.~R.}, \bibinfo{author}{Bilimoria, K.~Y.} \& \bibinfo{author}{Stulberg, J.~J.}
\newblock \bibinfo{journal}{\bibinfo{title}{Video-based feedback for the improvement of surgical technique: a platform for remote review and improvement of surgical technique}}.
\newblock {\emph{\JournalTitle{JAMA Surgery}}} \textbf{\bibinfo{volume}{155}}, \bibinfo{pages}{1078--1079} (\bibinfo{year}{2020}).

\bibitem{Yanik2024}
\bibinfo{author}{Yanik, E.}, \bibinfo{author}{Schwaitzberg, S.} \& \bibinfo{author}{De, S.}
\newblock \bibinfo{journal}{\bibinfo{title}{Deep learning for video-based assessment in surgery}}.
\newblock {\emph{\JournalTitle{JAMA Surgery}}} \textbf{\bibinfo{volume}{159}}, \bibinfo{pages}{957--958} (\bibinfo{year}{2024}).

\bibitem{Makary2013}
\bibinfo{author}{Makary, M.~A.}
\newblock \bibinfo{journal}{\bibinfo{title}{The power of video recording: taking quality to the next level}}.
\newblock {\emph{\JournalTitle{JAMA}}} \textbf{\bibinfo{volume}{309}}, \bibinfo{pages}{1591--1592} (\bibinfo{year}{2013}).

\bibitem{Boyle2025}
\bibinfo{author}{Boyle, C.}, \bibinfo{author}{Blackman, M.}, \bibinfo{author}{Hamilton, B.} \& \bibinfo{author}{Likosky, D.~S.}
\newblock \bibinfo{journal}{\bibinfo{title}{Applying elite tennis paradigms to surgical performance}}.
\newblock {\emph{\JournalTitle{JAMA Surgery}}}  (\bibinfo{year}{2025}).

\bibitem{Kudo2018}
\bibinfo{author}{Kudo, T.} \& \bibinfo{author}{Richardson, J.}
\newblock \bibinfo{journal}{\bibinfo{title}{Sentencepiece: A simple and language independent subword tokenizer and detokenizer for neural text processing}}.
\newblock {\emph{\JournalTitle{arXiv preprint arXiv:1808.06226}}}  (\bibinfo{year}{2018}).

\bibitem{Psychogyios2023}
\bibinfo{author}{Psychogyios, D.} \emph{et~al.}
\newblock \bibinfo{journal}{\bibinfo{title}{Sar-rarp50: Segmentation of surgical instrumentation and action recognition on robot-assisted radical prostatectomy challenge}}.
\newblock {\emph{\JournalTitle{arXiv preprint arXiv:2401.00496}}}  (\bibinfo{year}{2023}).

\bibitem{Lin2024}
\bibinfo{author}{Lin, B.} \emph{et~al.}
\newblock \bibinfo{title}{Video-llava: Learning united visual representation by alignment before projection}.
\newblock In \emph{\bibinfo{booktitle}{Proceedings of the 2024 conference on empirical methods in natural language processing}}, \bibinfo{pages}{5971--5984} (\bibinfo{year}{2024}).

\bibitem{Van2020}
\bibinfo{author}{van Amsterdam, B.}, \bibinfo{author}{Clarkson, M.~J.} \& \bibinfo{author}{Stoyanov, D.}
\newblock \bibinfo{title}{Multi-task recurrent neural network for surgical gesture recognition and progress prediction}.
\newblock In \emph{\bibinfo{booktitle}{2020 IEEE International Conference on Robotics and Automation (ICRA)}}, \bibinfo{pages}{1380--1386} (\bibinfo{organization}{IEEE}, \bibinfo{year}{2020}).

\end{thebibliography}
\end{document}